\documentclass[journal, 10pt, twocolumn]{IEEEtran}

\usepackage{cite}

\ifCLASSINFOpdf
\else
   \usepackage[dvips]{graphicx}
   \DeclareGraphicsExtensions{.eps}
\fi
\usepackage[cmex10]{amsmath}
\usepackage{algorithmic}

\usepackage{amssymb}

\usepackage{array}

\usepackage{mdwmath}
\usepackage{mdwtab}

\usepackage{eqparbox}

\usepackage{multirow}

\usepackage{fixltx2e}

\usepackage{enumerate}

\usepackage{url}

\usepackage{tabularx}

\begin{document}
%
\title{$\ell_0$ Sparsifying Transform Learning with Efficient Optimal Updates and Convergence Guarantees}


%
%
%

\author{Saiprasad~Ravishankar,~\IEEEmembership{Student Member,~IEEE,} and~Yoram~Bresler,~\IEEEmembership{Fellow,~IEEE}
\thanks{Copyright (c) 2014 IEEE. Personal use of this material is permitted. However, permission to use this material for any other purposes must be obtained from the IEEE by sending a request to pubs-permissions@ieee.org.}
\thanks{This work was first submitted to the IEEE Transactions on Signal Processing on April 02, 2014. It was accepted in the current form on January 09, 2015.}
\thanks{This work was supported in part by the National Science Foundation (NSF) under grants CCF-1018660 and CCF-1320953.}
\thanks{S. Ravishankar and Y. Bresler are with the Department of Electrical and Computer Engineering and the  Coordinated Science Laboratory, University of Illinois, Urbana-Champaign, IL, 61801 USA e-mail: (ravisha3, ybresler)@illinois.edu.}}
\maketitle

\begin{abstract}
Many applications in signal processing benefit from the sparsity of signals in a certain transform domain or dictionary. Synthesis sparsifying dictionaries that are directly adapted to data have been popular in applications such as image denoising, inpainting, and medical image reconstruction. In this work, we focus instead on the sparsifying transform model, and study the learning of well-conditioned square sparsifying transforms. The proposed algorithms alternate between a $\ell_0$ ``norm"-based sparse coding step, and a non-convex transform update step. We derive the exact analytical solution for each of these steps. The proposed solution for the transform update step achieves the global minimum in that step, and also provides speedups over iterative solutions involving conjugate gradients. 
We establish that our alternating algorithms are globally convergent to the set of local minimizers of the non-convex transform learning problems.
In practice, the algorithms are insensitive to initialization. We present results illustrating the promising performance and significant speed-ups of transform learning over synthesis K-SVD in image denoising.
\end{abstract}









%

\IEEEpeerreviewmaketitle

\section{Introduction} \label{introsec}

The sparsity of signals and images in a certain transform domain or dictionary has been widely exploited in numerous applications in recent years. While transforms are a classical tool in signal processing, alternative models have also been studied for sparse representation of data, most notably the popular \emph{synthesis model} \cite{elmiru, ambruck}, the \emph{analysis model}  \cite{elmiru} and its more realistic extension, the \emph{noisy signal analysis model} \cite{akd}. 
In this paper, we focus  specifically on the sparsifying \emph{transform model} \cite{tfcode, sabres}, which is a generalized analysis model, and suggests that a signal $y \in \mathbb{R}^{n}$ is \emph{approximately sparsifiable} using a transform $ W  \in \mathbb{R}^{m \times n} $, that is $Wy = x + e$ where $x \in \mathbb{R}^{m}$ is sparse in some sense, and $e$ is a small residual.
A distinguishing feature is that, unlike the synthesis or noisy signal analysis models, where the residual is measured in the signal domain, in the transform model the residual is in the transform domain. 

The transform model is not only more general in its modeling capabilities than the analysis models, it is also much more efficient and scalable than both the synthesis and noisy signal analysis models.
We briefly review the main distinctions between these sparse models (cf. \cite{sabres} for a more detailed review, and for the relevant references) in this and the following paragraphs.
One key difference is in the process of finding a sparse representation for data given the model, or dictionary. 
For the transform model, given the signal $y$ and transform $W$, the  \emph{transform sparse coding} problem  \cite{sabres} minimizes $\left \| Wy-x \right \|_{2}^{2}$ subject to  $\left \| x \right \|_{0}\leq s$, where $s$ is a given sparsity level. The solution $\hat{x}$ is obtained exactly and cheaply by zeroing out all but the $s$ coefficients of largest magnitude in $Wy$ \footnote{Moreover, given $W$ and sparse code $x$, we can also recover a least squares estimate of the underlying signal as $\hat{y} = W^{\dagger} x$, where $W^{\dagger}$ is the pseudo-inverse of $W$.}.
In contrast, for the synthesis or noisy analysis models, the process of \emph{sparse coding} is NP-hard (Non-deterministic Polynomial-time hard). 
While some of the approximate algorithms that have been proposed for synthesis or analysis sparse coding are guaranteed to provide the correct solution under certain conditions, in applications, especially those involving learning the models from data,  these conditions are often violated.
 Moreover, the various synthesis and analysis sparse coding algorithms tend to be computationally expensive for large-scale problems.

Recently, the data-driven adaptation of sparse models has received much interest.
The adaptation of synthesis dictionaries based on training signals \cite{ols, eng, Kruzdel, elad, Yagh, skret, Mai} has been shown to be  useful in various applications \cite{elad2, elad3, bresai}. The learning of analysis dictionaries, employing either the analysis model or its noisy signal extension, has also received some recent attention  \cite{nam, ophi, yll, akd}. 

Focusing instead on the transform model, we recently developed  the following formulation \cite{sabres} for the learning of well-conditioned square sparsifying transforms. Given a matrix $ Y  \in \mathbb{R}^{n \times N} $, whose columns represent training signals, our formulation for learning a square sparsifying transform $W \in \mathbb{R}^{n \times n}$ for $Y$ is \cite{sabres}
\begin{align} 
\nonumber (\mathrm{P0}) \;\; & \min_{W,X}\: \left \| WY-X \right \|_{F}^{2} + \lambda \, \begin{pmatrix}
 \xi \left \| W \right \|_{F}^{2} - \log \,\left | \mathrm{det \,} W \right |
\end{pmatrix} \\
\nonumber &\, s.t.\; \:  \left \| X _{i} \right \|_{0}\leq s\; \: \forall \,\,  i
\end{align}
where $\lambda>0$, $\xi >0$ are parameters, and $X  \in \mathbb{R}^{n \times N}$ is a matrix, whose columns $X _{i}$ are the sparse codes of the corresponding training signals $Y_i$. The term $\left \| WY-X \right \|_{F}^{2}$ in (P0) is called \emph{sparsification error}, and denotes the deviation of the data in the transform domain from its sparse approximation (i.e., the deviation of $WY$ from the sparse matrix $X$).
Problem (P0) also has $v(W) \triangleq - \log \,\left | \mathrm{det \,} W \right |  + \xi \left \| W \right \|_{F}^{2} $ as a regularizer in the objective to prevent trivial solutions.
We have proposed an alternating algorithm algorithm for solving (P0) \cite{sabres}, that alternates between updating $X$ (sparse coding), and $W$ (transform update), with the other variable kept fixed.

Because of the simplicity of sparse coding in the transform model, the alternating algorithm for transform learning  \cite{sabres} has a low computational cost. On the other hand, because, in the case of the synthesis or noisy signal analysis models, the learning formulations involve the NP-hard sparse coding, such learning formulations are also NP hard.  Moreover, even when the $\ell_0$ sparse coding is approximated by a convex relaxation, the various synthesis or analysis dictionary learning problems remain highly non-convex. 
Because the approximate algorithms for these problems usually solve the sparse coding problem repeatedly in the iterative process of adapting the sparse model, the cost of sparse coding is multiplied manyfold. Hence, the synthesis or analysis dictionary learning algorithms tend to be computationally expensive in practice for large scale problems.
Finally, popular algorithms for synthesis dictionary learning such as K-SVD \cite{elad}, or algorithms for analysis dictionary learning do not have convergence guarantees.

In this follow-on work on transform learning, keeping the focus on the square transform learning formulation (P0), we make the following contributions.
\begin{itemize}
\item We  derive highly \emph{efficient closed-form solutions} for the update steps in the alternating minimization procedure for (P0), that further enhance the computational properties of transform learning.
\item We also consider the alternating minimization of an alternative version of (P0) in this paper, that is obtained by replacing the sparsity constraints with sparsity penalties.
\item Importantly, we establish that our iterative algorithms for transform learning are globally convergent to the set of local minimizers of the non-convex transform learning problems.
\end{itemize}



We organize the rest of this paper as follows. Section \ref{sec2} briefly describes our transform learning formulations. In Section \ref{algo2}, we derive efficient algorithms for transform learning, and discuss the algorithms' computational cost. In Section \ref{convcomp}, we present convergence guarantees for our algorithms. The proof of convergence is provided in the Appendix. Section \ref{sec3} presents experimental results demonstrating the convergence behavior, and the computational efficiency of the proposed scheme. We also show brief results for the image denoising application. In Section \ref{sec4}, we conclude.




\section{Learning Formulations and Properties}
\label{sec2}

The transform learning Problem (P0) was introduced in Section \ref{introsec}. Here, we discuss some of its important properties.
The regularizer $v(W) \triangleq - \log \,\left | \mathrm{det \,} W \right |  + \xi \left \| W \right \|_{F}^{2} $ helps prevent trivial solutions in (P0).
The $ \log \,\left | \mathrm{det \,} W \right | $ penalty eliminates degenerate solutions such as those with zero, or repeated rows. 
While it is sufficient to consider the $\mathrm{det} \, W > 0$ case \cite{sabres}, to simplify the algorithmic derivation we replace the positivity constraint by the absolute value in the formulation in this paper.
The $\left \| W \right \|_{F}^{2}$ penalty in (P0) helps remove a `scale ambiguity' in the solution, which occurs when the data admit an exactly sparse representation \cite{sabres}.
The $- \log \,\left | \mathrm{det \,} W \right | $ and $\left \| W \right \|_{F}^{2}$ penalties together additionally help control the condition number $\kappa(W)$ of the learnt transform.
(Recall that the condition number of a matrix $A \in \mathbb{R}^{n \times n}$ is defined as $\kappa(A) = \beta_{1}/\beta_{n}$, where $\beta_{1}$ and $\beta_{n}$ denote the largest and smallest singular values of $A$, respectively.)
In particular, badly conditioned transforms typically convey little information and may degrade performance in applications such as signal/image representation, and denoising \cite{sabres}. Well-conditioned transforms, on the other hand, have been shown to perform well in (sparse) image representation, and denoising \cite{sabres, doubsp2l}.






The condition number $\kappa(W)$ can be upper bounded by a monotonically increasing function of $v(W)$ (see Proposition 1 of \cite{sabres}). Hence, minimizing $v(W)$ encourages reduction of the condition number. The regularizer $v(W)$ also penalizes bad scalings. Given a transform $W$ and a scalar $\alpha \in \mathbb{R}$, $v(\alpha W) \to \infty$ as the scaling $\alpha \to 0$ or $\alpha \to \infty$. For a fixed $\xi$, as $\lambda$ is increased in (P0), the optimal transform(s) become well-conditioned. In the limit $\lambda \to \infty$, their condition number tends to 1, and their spectral norm (or, scaling) tends to $1/\sqrt{2 \xi}$. Specifically, for $\xi=0.5$, as  $\lambda \to \infty$, the optimal transform tends to an \emph{orthonormal transform}. In practice, the transforms learnt via (P0) have condition numbers very close to 1 even for finite $\lambda$ \cite{sabres}. The specific choice of $\lambda$ depends on the application and desired condition number.



In this paper, to achieve invariance of the learned transform to trivial scaling of the training data $Y$, we set $\lambda = \lambda_{0} \left \| Y \right \|_{F}^{2}$ in (P0), where $\lambda_0>0$ is a constant. 
Indeed, when the data $Y$ are replaced with $\alpha Y$ ($\alpha \in \mathbb{R}$, $\alpha \neq 0$) in (P0), we can set $X=\alpha X'$. Then, the objective function becomes $\alpha^{2}\begin{pmatrix}
\left \| WY-X' \right \|_{F}^{2} +  \lambda_{0} \left \| Y \right \|_{F}^{2} \, v(W) 
\end{pmatrix}$, which is just a scaled version of the objective in (P0) (for un-scaled $Y$). Hence, its minimization over $(W, X')$ (with $X'$ constrained to have columns of sparsity $\leq s$) yields the same solution(s) as (P0). Thus, the learnt transform for data $\alpha Y$ is the same as for $Y$, while the learnt sparse code for $\alpha Y$ is $\alpha$ times that for $Y$.

We have shown \cite{sabres} that the cost function in (P0) is lower bounded by $\lambda v_{0}$, where $v_{0} =  \frac{n}{2} + \frac{n}{2} \log (2 \xi)$. The minimum objective value in Problem (P0) equals this lower bound if and only if there exists a pair $(\hat{W},\hat{X})$ such that $\hat{W}Y = \hat{X}$, with $\hat{X} \in \mathbb{R}^{n \times N}$ whose columns have sparsity $\leq s$, and $\hat{W} \in \mathbb{R}^{n \times n}$ whose singular values are all equal to $1/\sqrt{2 \xi}$ (hence, the condition number $\kappa (\hat{W})=1$). Thus, when an ``error-free" transform model exists for the data, and the underlying transform is unit conditioned, such a transform model is guaranteed to be a global minimizer of Problem (P0) (i.e., such a model is \emph{identifiable} by solving (P0)). Therefore, it makes sense to solve (P0) to find such good models.


Another interesting property of Problem (P0) is that it admits an equivalence class of solutions/minimizers. 
Because the objective in (P0) is unitarily invariant, then given a minimizer $(\tilde{W},\tilde{X})$, the pair $(\Theta \tilde{W}, \Theta \tilde{X})$ is another equivalent minimizer for all \emph{sparsity-preserving orthonormal matrices} $\Theta$, i.e., $\Theta$ such that $\| \Theta \tilde{X}_{i}  \|_{0} \leq s$ $\forall$ $i$. For example, $\Theta$ can be a row permutation matrix, or a diagonal $\pm 1$ sign matrix.



We note that a cost function similar to that in (P0), but lacking the $\left \| W \right \|_{F}^{2}$ penalty has been derived under certain assumptions in the very different setting of blind source separation \cite{mzib}. However, the transforms learnt via Problem (P0) perform poorly \cite{sabres} in signal processing applications, when the learning is done excluding the crucial $\left \| W \right \|_{F}^{2}$ penalty, which as discussed, helps overcome the scale ambiguity and control the condition number.

In this work, we also consider an alternative version of Problem (P0) by replacing
the $\ell_{0}$ sparsity constraints with $\ell_{0}$ penalties in the objective (this version of the transform learning problem has been recently used for example in adaptive tomographic reconstruction \cite{luke1, luke2}). In this case, we obtain the following unconstrained (or, sparsity penalized) transform learning problem
\begin{align} 
\nonumber (\mathrm{P1}) \;\; & \min_{W,X}\: \left \| WY-X \right \|_{F}^{2} + \lambda \, v(W) + \sum _{i=1}^{N} \eta_{i}^{2} \left \| X _{i} \right \|_{0} 
\end{align}
where $\eta_{i}^{2}$, with $\eta_{i} > 0$ $\forall \, i$, denote the weights (e.g., $\eta_{i} = \eta$ $\forall \, i$ for some $\eta$) for the sparsity penalties. The various aforementioned properties for Problem (P0) can be easily extended to the case of the alternative Problem (P1).




\section{Transform Learning Algorithm}  \label{algo2}

\subsection{Algorithm} \label{algo}

We have previously proposed \cite{sabres} an alternating algorithm for solving (P0) that alternates between solving for $X$ (\emph{sparse coding step}) and $W$ (\emph{transform update step}), with the other variable kept fixed. While the sparse coding step has an exact solution, the transform update step was performed using iterative nonlinear conjugate gradients (NLCG). This alternating algorithm for transform learning has a low computational cost compared to synthesis/analysis dictionary learning. In the following, we provide a further improvement: we show that both steps of transform learning (for either (P0) or (P1)) can in fact, be performed exactly and cheaply. 





\subsubsection{Sparse Coding Step} \label{algosparsecode1}
The sparse coding step in the alternating algorithm for (P0) is as follows \cite{sabres} 
\begin{equation} \label{z4}
 \min_{X}\: \left \| WY-X \right \|_{F}^{2}\;\: s.t.\; \:  \left \| X _{i} \right \|_{0}\leq s\; \: \forall \,\,  i
\end{equation}
The above problem is to project $WY$ onto the (non-convex) set of matrices whose columns have sparsity $\leq s$. Due to the additivity of the objective, this corresponds to projecting each column of $WY$ onto the set of sparse vectors $\left \{ x \in \mathbb{R}^{n} : \left \| x \right \|_{0} \leq s \right \}$, which we call the $s$-$\ell_{0}$ ball. Now, for  a vector $z \in \mathbb{R}^{n}$, the optimal projection $\hat{z}$ onto the $s$-$\ell_{0}$ ball is computed by zeroing out all but the $s$ coefficients of largest magnitude in $z$. If there is more than one choice for the $s$ coefficients of largest magnitude in $z$ (can occur when multiple entries in $z$ have identical magnitude), then the optimal $\hat{z}$ is not unique. We then choose $\hat{z} =  H_{s}(z)$, where $ H_{s}(z)$ is the projection, for which the indices of the $s$ largest magnitude elements (in $z$) are the lowest possible. Hence, an optimal sparse code in \eqref{z4} is computed as $\hat{X}_{i} =  H_{s}(W Y_{i})$  $\forall \, i$.

In the case of Problem (P1), we solve the following sparse coding problem
\begin{equation} \label{bbt5}
   \min_{X}\: \left \| WY-X \right \|_{F}^{2} +  \sum _{i=1}^{N} \eta_{i}^{2} \left \| X _{i} \right \|_{0}
\end{equation}
 A solution $\hat{X}$ of \eqref{bbt5} in this case is obtained as  $\hat{X}_{i} =  \hat{H}_{\eta_{i}}^{1} (W Y_{i})$ $\forall \, i$, where the (hard-thresholding) operator $\hat{H}_{\eta}^{1} (\cdot)$ is defined as follows.
\begin{equation} \label{equ88}
 \left ( \hat{H}_{\eta}^{1} (b) \right )_{j}=\left\{\begin{matrix}
 0&, \mathrm{if} \;\, \left | b_{j} \right | < \eta \\
b_{j}  & , \mathrm{if} \;\, \left | b_{j} \right | \geq \eta 
\end{matrix}\right.
\end{equation}
Here, $b \in \mathbb{R}^{n}$, and the subscript $j$ indexes vector entries. This form of the solution to \eqref{bbt5} has been mentioned in prior work \cite{luke1}. For completeness, we include a brief proof in Appendix \ref{appspcodepen}.
When the condition $\left | (WY)_{ji} \right | = \eta_{i}$ occurs for some $i$ and $j$ (where $(WY)_{ji}$ is the element of $WY$ on the $j^{\mathrm{th}}$ row and $i^{\mathrm{th}}$ column), the corresponding optimal $\hat{X}_{ji}$ in \eqref{bbt5}
can be either $(WY)_{ji} $ or $0$ (both of which correspond to the minimum value of the cost in \eqref{bbt5}). The definition in \eqref{equ88} breaks the tie between these equally valid solutions by selecting the first.
Thus, similar to Problem \eqref{z4}, the solution to \eqref{bbt5} can be computed exactly.

\newtheorem{thm}{Theorem}
\newtheorem{cor}{Corollary}
\newtheorem{lem}{Lemma}
\newtheorem{prop}{Proposition}

\subsubsection{Transform Update Step}
The transform update step of (P0) or (P1) involves the following unconstrained non-convex \cite{sabres} minimization.
\begin{equation} \label{z5}
\min_{W} \,\, \left \| WY-X \right \|_{F}^{2}+\lambda  \xi \left \| W \right \|_{F}^{2}- \lambda \log \,\left | \mathrm{det \,} W \right |
\end{equation}
Note that although NLCG works well for the transform update step \cite{sabres}, convergence to the global minimum of the non-convex transform update step has not been proved with NLCG. Instead, replacing NLCG, the following proposition provides the closed-form solution for Problem \eqref{z5}. The solution is written in terms of an appropriate singular value decomposition (SVD).
We use $( \cdot )^T$ to denote the matrix transpose operation, and $M^{\frac{1}{2}}$ to denote the positive definite square root of a positive definite matrix $M$. We let $I$ denote the $n \times n$ identity matrix. 




\begin{prop}\label{propel1} \vspace{0.02in}
Given the training data $Y \in \mathbb{R}^{n \times N}$, sparse code matrix $X \in \mathbb{R}^{n \times N}$,  and $\lambda>0$, $\xi>0$, factorize $ YY^{T} + \lambda \xi I$ as $LL^{T}$, with $L \in \mathbb{R}^{n \times n}$. Further, let $L^{-1}YX^{T}$ have a full SVD of $Q \Sigma R^{T}$. Then, a global minimizer for the transform update step \eqref{z5} can be written as 
\begin{equation} \label{tru1}
\hat{W}=0.5  R \left(\Sigma+ \left ( \Sigma^{2}+2\lambda I \right )^{\frac{1}{2}}\right)Q^{T}L^{-1}
\end{equation}
The solution is unique if and only if $L^{-1}YX^{T}$ is non-singular. Furthermore, the solution is invariant to the choice of factor $L$. 
\end{prop}


\hspace{0.1in} \textit{Proof:}   The objective function in \eqref{z5} can be re-written as $ tr \left \{ W\left ( YY^{T}+\lambda \xi I \right )W^{T} \right \}$ $- \,\, 2 \, tr (WYX^{T})  + tr (XX^{T}) $ $- \lambda \log \,\left | \mathrm{det \,} W \right | $. We then decompose the positive-definite matrix $ YY^{T} + \lambda \xi I$ as $LL^{T}$ (e.g.,  $L$ can be the positive-definite square root, or the cholesky factor of $ YY^{T} + \lambda \xi I$). The objective function then simplifies as follows
\[ tr \left ( WLL^{T}W^{T}-2WYX^{T}+XX^{T}  \right )- \lambda \log \,\left | \mathrm{det \,} W \right | \]
Using a change of variables $B = WL$, the multiplicativity of the determinant implies $\log \,\left | \mathrm{det \,} B \right | = \log \,\left | \mathrm{det \,} W \right | + \log \,\left | \mathrm{det \,} L \right |$. Problem \eqref{z5} is then equivalent to
\begin{equation} \label{z6}
\min_{B} \,\, tr\left ( BB^{T} \right ) - 2 tr\left ( BL^{-1}YX^{T} \right ) - \lambda \log \,\left | \mathrm{det \,} B \right |
\end{equation}

Next, let $B = U \Gamma V^{T}$, and $L^{-1}YX^{T}= Q \Sigma R^{T}$ be full SVDs ($U, \Gamma, V, Q, \Sigma, R$ are all $n \times n$ matrices), with $\gamma_{i}$ and $\sigma_{i}$ denoting the diagonal entries of $\Gamma$ and $\Sigma$, respectively. The unconstrained minimization \eqref{z6} then becomes
\[ \min_{\Gamma}\, \left [ tr\left ( \Gamma^{2} \right ) - 2 \, \max_{U, V} \left \{ tr\left (  U \Gamma V^{T}Q\Sigma R^{T}\right ) \right \} -\lambda \sum_{i=1}^{n} \log\,\gamma_{i} \right ] \]
For the inner maximization, we use the result $\max_{U, V} \, tr\left ( U \Gamma V^{T}Q\Sigma R^{T} \right )$ $=$ $tr\left ( \Gamma \Sigma \right )$ \cite{mirs}, where the maximum is attained by setting $U=R$ and $V=Q$. The remaining minimization with respect to $\Gamma$ is then
\begin{equation} \label{germa}
\min_{\{\gamma_{i}\}} \,\sum_{i=1}^{n} \gamma_{i}^{2}-2 \sum_{i=1}^{n}\gamma_{i} \sigma_{i} - \lambda \sum_{i=1}^{n} \log \,\gamma_{i}
\end{equation}
This problem is convex in the non-negative singular values $\gamma_{i}$, and the solution is obtained by differentiating the cost in \eqref{germa} with respect to the $\gamma_{i}$'s and setting the derivative to 0. This gives $\gamma_{i}= 0.5\begin{pmatrix}
\sigma_{i}\pm\sqrt{\sigma_{i}^{2}+2\lambda}
\end{pmatrix}$ $\forall \, i$. Since all the $\gamma_{i} \geq 0$, the only feasible solution is 
\begin{equation} \label{qdirfcwsffe}
 \gamma_{i}=\frac{\sigma_{i}+\sqrt{\sigma_{i}^{2}+2\lambda}}{2} \,\, \forall \, i 
\end{equation}
Thus, a closed-form solution or global minimizer for the transform update step \eqref{z5} is given as in \eqref{tru1}. 



The solution \eqref{tru1} is invariant to the specific choice of the matrix $L$.  To show this, we will first show that if $L_{1} \in \mathbb{R}^{n \times n}$ and $L_{2} \in \mathbb{R}^{n \times n}$ satisfy $ YY^{T} + \lambda \xi I=L_{1}L_{1}^{T}=L_{2}L_{2}^{T}$, then $L_{2}=L_{1}G$, where $G$ is an orthonormal matrix satisfying $G G^{T} = I$. A brief proof of the latter result is as follows. 
Since $L_1$ and $L_2$ are both $n \times n$ full rank matrices (being square roots of the positive definite matrix $ YY^{T} + \lambda \xi I$), we have $L_2 = L_1 \begin{pmatrix} L_{1}^{-1}L_{2}\end{pmatrix}= L_{1}G$, with $G \triangleq L_{1}^{-1}L_{2}$ a full rank matrix. Moreover, since $L_{1}L_{1}^{T} - L_{2}L_{2}^{T} = 0$, we have
\begin{equation} \label{qdoed2s}
L_{1}\begin{pmatrix}
I -  G G^{T}
\end{pmatrix}L_{1}^{T} = 0
\end{equation}
Because  $L_1$ has full rank, we must therefore have that $GG^{T} = I$ for \eqref{qdoed2s} to hold. Therefore, $G$ is an orthonormal matrix satisfying $G G^{T} = I$.

Consider $L_1$ and $L_2$ as defined above. Now, if $Q_{1}$ is the left singular matrix corresponding to $L_{1}^{-1}Y X^{T}$, then $Q_{2} = G^{T} Q_{1}$ is a corresponding left singular matrix for $L_{2}^{-1} YX^{T} = G^{T} L_{1}^{-1} YX^{T}$. 
Therefore, replacing $L_1$ by $L_2$  in \eqref{tru1}, making the substitutions  $L_{2}^{-1} = G^{T} L_{1}^{-1}$, $Q_{2}^{T} = Q_{1}^{T}  G$, and using the orthonormality of $G$, it is obvious that the closed-form solution  \eqref{tru1} involving $L_2$ is identical to that involving $L_1$.


Finally, we show that the solution \eqref{tru1} is unique if and only if $L^{-1}YX^{T}$ is non-singular.
First, the solution \eqref{tru1} can be written using the notations introduced above as $\hat{W} = \begin{pmatrix}
\sum_{i=1}^{n} \gamma_{i} R_{i}Q_{i}^{T}
\end{pmatrix}L^{-1}$, where $R_i$ and $Q_i$ are the $\mathrm{i^{th}}$ columns of $R$ and $Q$, respectively. We first show that the non-singularity of $L^{-1}YX^{T}$ is a necessary condition for uniqueness of the solution to  \eqref{z5}.
Now, if $L^{-1}YX^{T}$ has rank $< n$, then a singular vector pair $\begin{pmatrix}
Q_{k}, R_{k}
\end{pmatrix}$ of $L^{-1}YX^{T}$ corresponding to a zero singular value (say $\sigma_k=0$) can also be modified as   $\begin{pmatrix}
Q_{k}, -R_{k}
\end{pmatrix}$ or  $\begin{pmatrix}
-Q_{k}, R_{k}
\end{pmatrix}$, yielding equally valid alternative SVDs of $L^{-1}YX^{T}$. However, because zero singular values in the matrix $\Sigma$ are mapped to non-zero singular values in the matrix $\Gamma$ (by \eqref{qdirfcwsffe}), we have that the following two matrices are equally valid solutions to  \eqref{z5}.
\begin{align}
\hat{W}^{a} = & \begin{pmatrix}
 \sum_{i\neq k} \gamma_{i} R_{i}Q_{i}^{T} + \gamma_{k} R_{k}Q_{k}^{T}
\end{pmatrix}L^{-1} \\
\hat{W}^{b} = & \begin{pmatrix}
 \sum_{i\neq k} \gamma_{i} R_{i}Q_{i}^{T} - \gamma_{k} R_{k}Q_{k}^{T}
\end{pmatrix}L^{-1} 
\end{align}
where $\gamma_k>0$. It is obvious that $\hat{W}^{a} \neq \hat{W}^{b}$, i.e., the optimal transform is not unique in this case. Therefore, $L^{-1}YX^{T}$ being non-singular is a necessary condition for uniqueness of the solution to  \eqref{z5}. 

Next, we show that the non-singularity of $L^{-1}YX^{T}$ is also a sufficient condition for the aforementioned uniqueness. First, if the singular values of  $L^{-1}YX^{T}$ are non-degenerate (distinct and non-zero), then the SVD of $L^{-1}YX^{T}$ is unique up to joint scaling of any pair $\begin{pmatrix}
Q_{i}, R_{i}
\end{pmatrix}$ by $\pm 1$. This immediately implies that the solution $\hat{W} = \begin{pmatrix}
\sum_{i=1}^{n} \gamma_{i} R_{i}Q_{i}^{T}
\end{pmatrix}L^{-1}$ is unique in this case. On the other hand, if  $L^{-1}YX^{T}$ has some repeated but still non-zero singular values, then they are mapped to repeated (and non-zero) singular values in $\Gamma$ by \eqref{qdirfcwsffe}. Let us assume that $\Sigma$ has only one singular value that repeats (the proof easily extends to the case of multiple repeated singular values) say $r$ times, and that these repeated values are arranged in the bottom half of the matrix $\Sigma$ (i.e., $\sigma_{n-r+1}= \sigma_{n-r+2}=...=\sigma_{n} = \hat{\sigma}>0$). Then, we have
\begin{equation} \label{eer345}
L^{-1}YX^{T} = \sum_{i=1}^{n-r} \sigma_{i} Q_{i}R_{i}^{T} + \hat{\sigma}\begin{pmatrix}
\sum_{i=n-r+1}^{n} Q_{i}R_{i}^{T}
\end{pmatrix}
\end{equation}
Because the matrix defined by the first sum on the right (in \eqref{eer345}) corresponding to distinct singular values is unique, and $\hat{\sigma}>0$, so too is the second matrix defined by the second sum.
(This is also a  simple consequence of the fact that although the singular vectors associated wtih repeated singular values are not unique, the subspaces spanned by them are.)
The transform update solution  \eqref{tru1} in this case is given as
\begin{equation} \label{fefvfev}
\hat{W} =  \begin{Bmatrix}
 \sum_{i=1}^{n-r} \gamma_{i} R_{i}Q_{i}^{T} + \gamma_{n-r+1} \begin{pmatrix}
\sum_{i=n-r+1}^{n} R_{i}Q_{i}^{T}
\end{pmatrix}
\end{Bmatrix}L^{-1}
\end{equation}
Based on the preceding arguments, it is clear that the right hand side of \eqref{fefvfev} is unique, irrespective of the particular choice of (non-unique) $Q$ and $R$. 
Thus, the transform update solution  \eqref{tru1} is unique when $L^{-1}YX^{T}$ has possibly repeated, but non-zero singular values.
Therefore, the non-singularity of $L^{-1}YX^{T}$ is also a sufficient condition for the  uniqueness of the solution to  \eqref{z5}.
$\;\;\; \blacksquare$

The transform update solution \eqref{tru1} is expressed in terms of the full SVD of  $L^{-1}YX^{T}$, where $L$ is for example, the Cholesky factor, or alternatively, the Eigenvalue Decomposition (EVD) square root of $YY^{T} + \lambda \xi I$. Although in practice the SVD, or even the square root of non-negative scalars, are computed using iterative methods, we will assume in the theoretical analysis in this paper, that the solution \eqref{tru1} is computed exactly. 
In practice, standard numerical methods are guaranteed to quickly provide machine precision accuracy for the SVD and other computations. Therefore, the transform update solution \eqref{tru1} is computed to within machine precision accuracy in practice.



The algorithms A1 and A2 (corresponding to Problems (P0) and (P1), respectively) for transform learning are shown in Fig. \ref{im5p}.

While Proposition \ref{propel1} provides the closed-form solution to equation \eqref{z5} for real-valued matrices, the solution can be extended to the complex-valued case (useful in applications such as magnetic resonance imaging (MRI) \cite{bresai}) by replacing the $(\cdot)^{T}$ operation in Proposition \ref{propel1} and its proof by $(\cdot)^{H}$, the Hermitian transpose operation. The same proof applies, with the trace maximization result for the real case replaced by $\max_{U, V} \, Re\left \{ tr\left ( U \Gamma V^{H}Q\Sigma R^{H} \right ) \right \} $ $=$ $tr\left ( \Gamma \Sigma \right )$ for the complex case, where $Re(A)$ denotes the real part of scalar $A$.



\begin{figure}
\begin{tabular}{p{8.3cm}}
\hline
Transform Learning Algorithms A1 and A2\\
\hline
 \textbf{Input\;:} \:\:\: $ Y $ - training data, $s$ - sparsity, $\lambda$ - constant, $\xi$ - constant, $J_0$ - number of iterations.\\
 \textbf{Output\;:} \:\:\: $\hat{W}$ - learned transform, $\hat{X}$ - learned sparse code matrix. \\
\textbf{Initial Estimates:} $(W^{0}, X^{0})$. \\
\textbf{Pre-Compute:} $L^{-1} = \left ( YY^{T} + \lambda \xi I \right )^{-1/2}$. \\
\textbf{For \;k = 1: $\mathbf{J_0}$ Repeat}\\
\begin{enumerate}
\item Compute full SVD of $L^{-1}Y ( \hat{X}^{k-1}  )^{T}$ as $Q \Sigma R^{T}$.
\item $\hat{W}^{k} = 0.5  R \left(\Sigma+ \left ( \Sigma^{2}+2\lambda I \right )^{\frac{1}{2}}\right)Q^{T}L^{-1}$.
\item $\hat{X}_{i}^{k} =  H_{s}(W^{k} Y_{i})$  $\forall \, i$ for Algorithm A1, or $\hat{X}_{i}^{k} =  \hat{H}_{\eta_{i}}^{1} (W^{k} Y_{i})$ $\forall \, i$ for Algorithm A2.
\end{enumerate}\\
\textbf{End} \\
\hline
\end{tabular}
\caption{Algorithms A1 and A2 for solving Problems (P0) and (P1), respectively. A superscript of $k$ is used to denote the iterates in the algorithms. Although we begin with the transform update step in each iteration above, one could alternatively start with the sparse coding step as well.} \label{im5p}
\end{figure}




\subsection{The Orthonormal Transform Limit} \label{limitalg}

We have seen that for $\xi=0.5$, as $\lambda \to \infty$, the $W$ minimizing (P0) tends to an orthonormal matrix. Here, we study the behavior of the actual sparse coding and transform update steps of our algorithm as the parameter $\lambda$ (or, equivalently $\lambda_0$, since $\lambda = \lambda_{0} \left \| Y \right \|_{F}^{2}$) tends to infinity. 
The following Proposition \ref{propel33} establishes that as $\lambda \to \infty$ with $\xi$ held at $0.5$, the sparse coding and transform update solutions for (P0) approach the corresponding solutions for an orthonormal transform learning problem. Although we consider Problem (P0) here, a similar result also holds with respect to Problem (P1).


\begin{prop}\label{propel33} \vspace{0.02in}
For $\xi = 0.5$, as $\lambda \to \infty$,  the sparse coding and transform update solutions in (P0) coincide with the corresponding solutions obtained by employing alternating minimization on the following orthonormal transform learning problem. 
\begin{equation} \label{orthprobav}
\min_{W,X}\:  \left \| WY-X \right \|_{F}^{2} \;\;   s.t.\; \: W^{T}W= I,\;  \left \| X _{i} \right \|_{0}\leq s\; \: \forall \,\,  i
\end{equation}
Specifically, the sparse coding step for Problem \eqref{orthprobav}  involves
\begin{equation*}
 \min_{X}\: \left \| WY-X \right \|_{F}^{2}\;\: s.t.\; \:  \left \| X _{i} \right \|_{0}\leq s\; \: \forall \,\,  i
\end{equation*}
and the solution is $\hat{X}_{i} =  H_{s}(W Y_{i})$  $\forall \, i$. Moreover, the transform update step for Problem \eqref{orthprobav} involves
\begin{equation} \label{ortp}
 \max_{W} \,tr\left ( W Y X^{T} \right ) \;\: s.t.\; \: W^{T} W=  I
\end{equation}
Denoting the full SVD of $YX^{T}$ by $U \Sigma V^{T}$, where $U \in \mathbb{R}^{n \times n}$, $\Sigma \in \mathbb{R}^{n \times n}$, $V \in \mathbb{R}^{n \times n}$, the optimal solution in Problem \eqref{ortp} is $\hat{W}=VU^{T}$. This solution is unique if and only if the singular values of $YX^{T}$ are non-zero.
\end{prop} \vspace{0.09in}

For $\xi \neq 0.5$, Proposition \ref{propel33} holds with  the constraint $W^{T}W= I$ in Problem \eqref{orthprobav} replaced by the constraint $W^{T}W= (1/2 \xi) I$. The transform update solution for  Problem \eqref{orthprobav} with the modified constraint  $W^{T}W= (1/2 \xi) I$ is the same as mentioned in Proposition \ref{propel33}, except for an additional scaling of $1/\sqrt{2 \xi}$.
The proof of  Proposition \ref{propel33} is provided in Appendix \ref{orthlimproof}.

The orthonormal transform case is special, in that Problem \eqref{orthprobav} is also an orthonormal
synthesis dictionary learning problem, with $W^{T}$ denoting the synthesis dictionary. This follows  immediately, using the identity $\|WY-X \|_F = \|Y - W^TX\|_F$, for orthonormal $W$. Hence, Proposition \ref{propel33} provides an alternating algorithm with optimal updates not only for the orthonormal transform learning problem, but at the same time for the orthonormal dictionary learning problem.

\subsection{Computational Cost} \label{compco}
The proposed transform learning algorithms A1 and A2 alternate between the sparse coding and transform update steps. Each of these steps has a closed-form solution. We now discuss their computational costs. We assume that the matrices $YY^{T} + \lambda I$ and $L^{-1}$ (used in \eqref{tru1}) are pre-computed (at the beginning of the algorithm) at total costs of $O(Nn^2)$ and $O(n^{3})$, respectively, for the entire algorithm. 


The computational cost of the sparse coding step in both Algorithms A1 and A2 is dominated by the computation of the product $WY$, and therefore scales as $ O(Nn^{2}) $. In contrast, the projection onto the $s$-$\ell_{0}$ ball in Algorithm A1 requires only $ O(nN \log n) $ operations, when employing sorting \cite{sabres}, and the hard thresholding (as in equation \eqref{equ88}) in Algorithm A2 requires only $O(nN)$ comparisons.

For the transform update step, the computation of the product $YX^{T}$ requires $\alpha N n^{2}$ multiply-add operations for an $X$ with $s$-sparse columns, and $s=\alpha n$.  Then, the computation of $L^{-1}Y X^{T}$, its SVD, and the closed-form transform update \eqref{tru1} require $O(n^{3})$ operations. On the other hand, when NLCG is employed for transform update, the cost (excluding the $YX^{T}$ pre-computation) scales as $ O(J n^{3}) $, where $J$ is the number of NLCG iterations \cite{sabres}. Thus, compared to NLCG, the proposed update formula \eqref{tru1} allows for both an exact and potentially cheap (depending on $J$) solution to the transform update step.


Under the assumption that $n \ll N$, the total cost per iteration (of sparse coding and transform update) of the proposed algorithms scales as $O(N n^{2})$. This is much lower than the per-iteration cost of learning an $n \times K$  overcomplete ($K>n$) synthesis dictionary $D$ using K-SVD \cite{elad}, which scales (assuming that the synthesis sparsity level $s \propto n$) as $O(KNn^{2})$. Our transform learning schemes also hold a similar computational advantage over analysis dictionary learning schemes such as analysis K-SVD \cite{sabres}.

As illustrated in Section \ref{convb}, our algorithms converge in few iterations in practice. Therefore, the per-iteration computational advantages (e.g., over K-SVD) also typically translate to a net computational advantage in practice (e.g., in denoising).

\section{Main Convergence Results} \label{convcomp}



\subsection{Result for Problem (P0)}



Problem (P0) has the constraint $\left \| X _{i} \right \|_{0}\leq s\, \forall \, i$, which can instead  be added as a penalty in the objective by using a barrier function $\psi (X)$ (which takes the value $+ \infty$ when the constraint is violated, and is zero otherwise). In this form, Problem (P0) is unconstrained, and we denote its objective as $g(W, X) =  \left \| WY-X \right \|_{F}^{2} + \lambda \xi \left \| W \right \|_{F}^{2}$ $ - \lambda \log \,\left | \mathrm{det \,} W \right | + \psi(X)$. The unconstrained minimization problem involving the objective $g(W, X)$ is exactly equivalent to the constrained formulation (P0) in the sense that the minimum objective values as well as the set of minimizers of the two formulations are identical. 
To see this, note that whenever the constraint $\left \| X _{i} \right \|_{0}\leq s\, \forall \, i$ is satisfied, the two objectives coincide.
Otherwise, the objective in the unconstrained formulation takes the value $+\infty$ and therefore, its minimum value is achieved where the constraint $\left \| X _{i} \right \|_{0}\leq s\, \forall \, i$  holds. This minimum value (and the corresponding set of minimizers) is therefore the same as that for the constrained formulation (P0). The proposed Algorithm A1 is an exact alternating algorithm for both the constrained and unconstrained formulations above.

 Problem (P0) is to find the best possible transform model for the given training data $Y$ by minimizing the sparsification error, and controlling the condition number (avoiding triviality). We are interested to know whether the proposed alternating algorithm converges to a minimizer of (P0), or whether it could get stuck in saddle points, or some non-stationary points. Problem (P0) is non-convex, and therefore well-known results on convergence of alternating minimization (e.g., \cite{tseng65}) do not apply here.
 The following Theorem \ref{theorem2} provides the convergence of our Algorithm A1 for (P0).  
We say that a sequence $ \left \{ a^{k} \right \}$ has an accumulation point $a$, if there is a subsequence that converges to $a$.
For a vector $h$, we let $\phi_{j}(h)$ denote the magnitude of the $j^{\mathrm{th}}$ largest element (magnitude-wise) of $h$. For some matrix $B$, $\left \| B \right \|_{\infty} \triangleq \max_{i,j} \left | B_{ij} \right |$.

\begin{thm}\label{theorem2}
Let $ \left \{ W^{k}, X^{k} \right \}$ denote the iterate sequence generated by Algorithm A1 with training data $Y$ and initial $(W^{0}, X^{0})$. Then, the objective sequence $ \left \{ g(W^{k}, X^{k}) \right \}$ is monotone decreasing, and converges to a finite value, say $g^{*}= g^{*}\left ( W^{0},X^{0} \right )$.
Moreover, the iterate sequence is bounded, and each accumulation point $(W, X)$ of the iterate sequence is a fixed point of the algorithm, and a local minimizer of the objective $g$ in the following sense. For each accumulation point $(W, X)$, $\exists$ $\epsilon' = \epsilon'(W) >0$ such that
\begin{equation} \label{dde}
g(W+dW, X+\Delta X) \geq g(W, X) =  g^{*}
\end{equation}
holds $\forall$ $dW \in \mathbb{R}^{n \times n}$ satisfying $\left \| dW \right \|_{F} \leq \epsilon'$, and all $\Delta X \in \mathbb{R}^{n \times N}$ in the union of the following regions.
\begin{itemize}
\item[R1.] The half-space $tr\left \{ (WY-X)\Delta X^{T} \right \} \leq 0$.  
\item[R2.] The local region defined by \\
$\left \| \Delta X \right \|_{\infty} < \min_{i}\left \{ \phi_{s}(W Y_{i}) : \left \| W Y_{i} \right \|_{0}>s \right \}$.
\end{itemize}
 Furthermore, if we have $ \left \|  W Y_{i} \right \|_{0} \leq s \, \forall \, i$, then $\Delta X$ can be arbitrary. 
\end{thm}
\vspace{0.07in}



The notation $g^{*}\left ( W^{0},X^{0} \right )$ in Theorem \ref{theorem2} represents the value to which the objective sequence $ \left \{ g(W^{k}, X^{k}) \right \}$ converges, starting from  an initial (estimate) $(W^{0}, X^{0})$.
Local region R2 in Theorem \ref{theorem2}  is defined in terms of the scalar $\min_{i}\left \{ \phi_{s}(W Y_{i}) : \left \| W Y_{i} \right \|_{0}>s \right \}$, which is computed by taking the columns of $WY$ with sparsity greater than $s$, and finding the $s$-largest magnitude element in each of these columns, and choosing the smallest of those magnitudes. The intuition for this particular construction of the local region is provided in the proof of Lemma \ref{lemma3c} in Appendix \ref{conproof}.

Theorem \ref{theorem2} indicates local convergence of our alternating Algorithm A1.
Assuming a particular initial $(W^{0}, X^{0})$, we have that every accumulation point $(W, X)$ of the iterate sequence is a local optimum by equation \eqref{dde}, and satisfies $g(W, X) =  g^{*}\left ( W^{0},X^{0} \right )$. Thus, all accumulation points of the iterates (for a particular initial $(W^{0}, X^{0})$) are equivalent (in terms of their cost), or are equally good local minima. We thus have the following corollary.

\begin{figure}[!t]
\begin{center}
\begin{tabular}{c}
\includegraphics[height=1.24in]{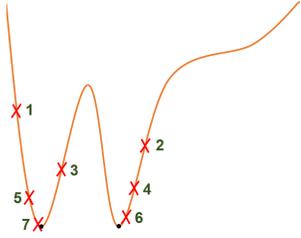}\\
\end{tabular}
\caption{Possible behavior of the algorithm near two hypothetical local minima (marked with black dots) of the objective. The numbered iterate sequence here has two subsequences (one even numbered, and one odd numbered) that converge to the two equally good (i.e., corresponding to the same value of the objective) local minima.}
\label{im10we}
\end{center}
\end{figure}

\begin{cor}\label{corollary1}
For Algorithm A1, assuming a particular initial $(W^{0}, X^{0})$, the objective converges to a local minimum, and the iterates converge to an equivalence class of local minimizers. 
\end{cor}
\vspace{0.04in}
The local optimality condition \eqref{dde} holds for the algorithm irrespective of initialization. However, the local minimum $ g^{*}\left ( W^{0},X^{0} \right )$ that the objective converges to may possibly depend on (i.e., vary with) initialization. Nonetheless, empirical evidence presented in Section \ref{sec3} suggests that the proposed transform learning scheme is insensitive to initialization. This leads us to conjecture that our algorithm could potentially converge to the global minimizer(s) of the learning problem in some (practical) scenarios. Fig. \ref{im10we} provides a simple illustration of the convergence behavior of our algorithm. We also have the following corollary of Theorem \ref{theorem2}, where `globally convergent' refers to convergence from any initialization.

\begin{cor}\label{corollary23}
 Algorithm A1 is globally convergent to the set of local minimizers of the non-convex transform learning objective $g(W, X)$.
\end{cor}


Note that our convergence result for the proposed non-convex learning algorithm, is free of any extra conditions or requirements. This is in clear distinction to algorithms such as IHT \cite{ihtblu, ihtloc} that solve non-convex problems, but require extra stringent conditions (e.g., tight conditions on restricted isometry constants of certain matrices) for their convergence results to hold. Theorem \ref{theorem2} also holds for any choice of the parameter $\lambda_0$ (or, equivalently $\lambda$) in (P0), that controls the condition number.

The optimality condition \eqref{dde} in Theorem \ref{theorem2} holds true not only for local (small) perturbations in $X$, but also for arbitrarily large perturbations of $X$ in a half space. For a particular accumulation point $(W, X)$, the condition $tr\left \{ (WY-X)\Delta X^{T} \right \} \leq 0$ in Theorem \ref{theorem2} defines a half-space of permissible perturbations in $\mathbb{R}^{n \times N}$. Now, even among the perturbations outside this half-space, i.e., $\Delta X$ satisfying $tr\left \{ (WY-X)\Delta X^{T} \right \} > 0$ (and also outside the local region R2 in Theorem \ref{theorem2}), we only need to be concerned about the perturbations that maintain the sparsity level, i.e., $\Delta X$ such that $X + \Delta X$ has sparsity $\leq s$ per column. For any other $\Delta X$, $g(W+ dW, X+ \Delta X) = + \infty > g(W, X)$ trivially. Now, since $X + \Delta X$ needs to have sparsity $\leq s$ per column, $\Delta X$ itself can be at most $2s$ sparse per column. Therefore, the condition $tr\left \{ (WY-X)\Delta X^{T} \right \} > 0$ (corresponding to perturbations that could violate \eqref{dde}) essentially corresponds to a union of low dimensional half-spaces (each corresponding to a different possible choice of support of $\Delta X$). In other words, the set of ``bad" perturbations is vanishingly small in $\mathbb{R}^{n \times N}$.


Note that Problem (P0) can be directly used for adaptive sparse representation (compression) of images \cite{sabres, yblsgg}, in which case the convergence results here are directly applicable. (P0) can also be used in applications such as blind denoising \cite{doubsp2l}, and blind compressed sensing \cite{syber}. The overall problem formulations \cite{doubsp2l, syber} in these applications are highly non-convex (see Section \ref{imden}). However, the problems are solved using alternating optimization \cite{doubsp2l, syber}, and the transform learning Problem (P0) arises as a sub-problem. Therefore, by using the proposed learning scheme, the transform learning step of the alternating algorithms for denoising/compressed sensing can be guaranteed (by Theorem \ref{theorem2}) to converge \footnote{Even when different columns of $X$ are required to have different sparsity levels in (P0), our learning algorithm and Theorem \ref{theorem2} can be trivially modified to guarantee convergence.}.

\subsection{Result for Penalized Problem (P1)}

When the sparsity constraints in (P0) are replaced with $\ell_{0}$ penalties in the objective with weights $\eta_{i}^{2}$ (i.e., we solve Problem (P1)), we obtain an unconstrained transform learning problem with objective $u(W, X) =  \left \| WY-X \right \|_{F}^{2} + \lambda \xi \left \| W \right \|_{F}^{2}$ $ - \lambda \log \,\left | \mathrm{det \,} W \right | + \sum _{i=1}^{N} \eta_{i}^{2} \left \| X _{i} \right \|_{0}$. In this case too, we have a convergence guarantee (similar to Theorem \ref{theorem2}) for Algorithm A2 that minimizes $u(W, X)$.

\begin{thm}\label{theorem3}
Let $ \left \{ W^{k}, X^{k} \right \}$ denote the iterate sequence generated by Algorithm A2 with training data $Y$ and initial $(W^{0}, X^{0})$. Then, the objective sequence $ \left \{ u(W^{k}, X^{k}) \right \}$ is monotone decreasing, and converges to a finite value, say $u^{*}= u^{*}\left ( W^{0},X^{0} \right )$.
Moreover, the iterate sequence is bounded, and each accumulation point $(W, X)$ of the iterate sequence is a fixed point of the algorithm, and a local minimizer of the objective $u$ in the following sense. For each accumulation point $(W, X)$, $\exists$ $\epsilon' = \epsilon'(W) >0$ such that
\begin{equation} \label{dde2}
u(W+dW, X+\Delta X) \geq u(W, X) =  u^{*}
\end{equation}
holds $\forall$ $dW \in \mathbb{R}^{n \times n}$ satisfying $\left \| dW \right \|_{F} \leq \epsilon'$, and all $\Delta X \in \mathbb{R}^{n \times N}$ satisfying $\left \| \Delta X \right \|_{\infty} < \min_{i}\left \{ \eta_{i}/2 \right \}$.  
\end{thm}
\vspace{0.07in}


The proofs of Theorems \ref{theorem2} and \ref{theorem3} are provided in Appendix \ref{conproof} and \ref{app5}.
Owing to Theorem 2, results analogous to Corollaries \ref{corollary1} and \ref{corollary23} apply.

\begin{cor}\label{corollary3bb}
Corollaries \ref{corollary1} and \ref{corollary23} apply to Algorithm A2 and the corresponding objective $u(W, X)$ as well.
\end{cor}

\section{Experiments}
\label{sec3}


\subsection{Framework} \label{expframew}

In this section, we present results demonstrating the properties of our proposed transform learning Algorithm A1 for (P0), and its usefulness in applications. First, we illustrate the convergence behavior of our alternating learning algorithm. We consider various initializations for transform learning and investigate whether the proposed algorithm is sensitive to initializations. This  study will provide some (limited) empirical understanding of local/global convergence behavior of the algorithm. Then, we compare our proposed algorithm to the NLCG-based transform learning algorithm \cite{sabres} at various patch sizes, in terms of image representation quality and computational cost of learning. Finally, we briefly discuss the usefulness of the proposed scheme in image denoising.

All our implementations were coded in Matlab version R2013a.  All computations were performed with an Intel Core i5 CPU at 2.5GHz and 4GB memory, employing a 64-bit Windows 7 operating system.

The data in our experiments are generated as the 2D patches of natural images. We use our transform learning Problem (P0) to learn adaptive sparse representations of such image patches. The means of the patches are removed and we only sparsify the mean-subtracted patches which are stacked as columns of the training matrix $Y$ (patches reshaped as vectors) in (P0). The means are added back for image display. Mean removal is typically adopted in image processing applications such as compression and denoising \cite{el2, doubsp2l}. Similar to prior work \cite{sabres, doubsp2l}, the weight $\xi=1$ in all our experiments.

We have previously introduced several metrics to evaluate the quality of learnt transforms \cite{sabres, doubsp2l}. The \emph{normalized sparsification error} (NSE) for a transform $W$ is defined as $ \| WY-X \|_{F}^{2}/$ $ \| WY  \|_{F}^{2}$, where $Y$ is the data matrix, and the columns $X_{i}=H_{s}(WY_{i})$ of the matrix $X$ denote the sparse codes \cite{sabres}. The NSE measures the fraction of energy lost in sparse fitting in the transform domain, and is an interesting property to observe for the learnt transforms. A useful performance metric for learnt transforms in image representation is the recovery peak signal to noise ratio (or \emph{recovery PSNR}), which was previously defined as $255 \sqrt{P}/$ $\left \| Y-W^{-1}X \right \|_{F}$ in decibels (dB), where $P$ is the number of image pixels and $X$ is again the transform sparse code of data $Y$ \cite{sabres}. The recovery PSNR measures the error in recovering the patches $Y$ (or equivalently the image, in the case of non-overlapping patches) as $W^{-1}X$ from their sparse codes $X$. The recovery PSNR serves as a simple surrogate for the performance of the learnt transform in compression. Note that if the proposed approach were to be used for compression, then the  $W$ matrix too would have to be transmitted as side information.

\subsection{Convergence Behavior} \label{convb}

Here, we study the convergence behavior of the proposed transform learning Algorithm A1. We extract the $8 \times 8$ ($n=64$) non-overlapping (mean-subtracted) patches of the $512 \times 512$ image Barbara \cite{elad}. Problem (P0) is solved to learn a square transform $W$ that is adapted to this data.
The data matrix $Y$ in this case has $N = 4096$ training signals (patches represented as vectors). The parameters are set as $s=11$, $\lambda_{0} = 3.1 \times 10^{-3}$. The choice of $\lambda_0$ here ensures well-conditioning of the learnt transform. Badly conditioned transforms degrade performance in applications \cite{sabres, doubsp2l}. Hence, we focus our investigation here only on the well-conditioned scenario.

\begin{figure}[!t]
\begin{center}
\begin{tabular}{cc}
\includegraphics[height=1.38in]{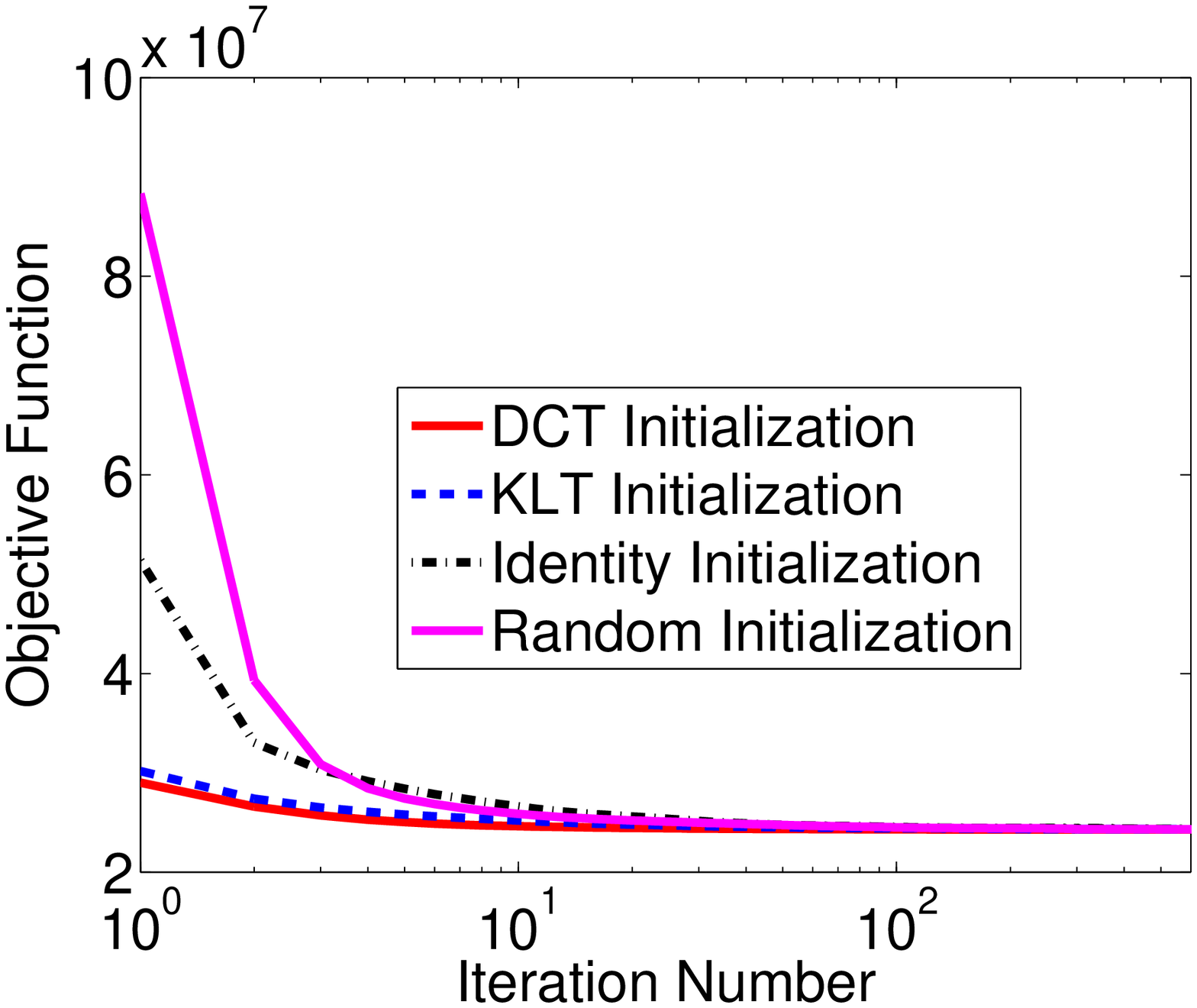}&
\includegraphics[height=1.38in]{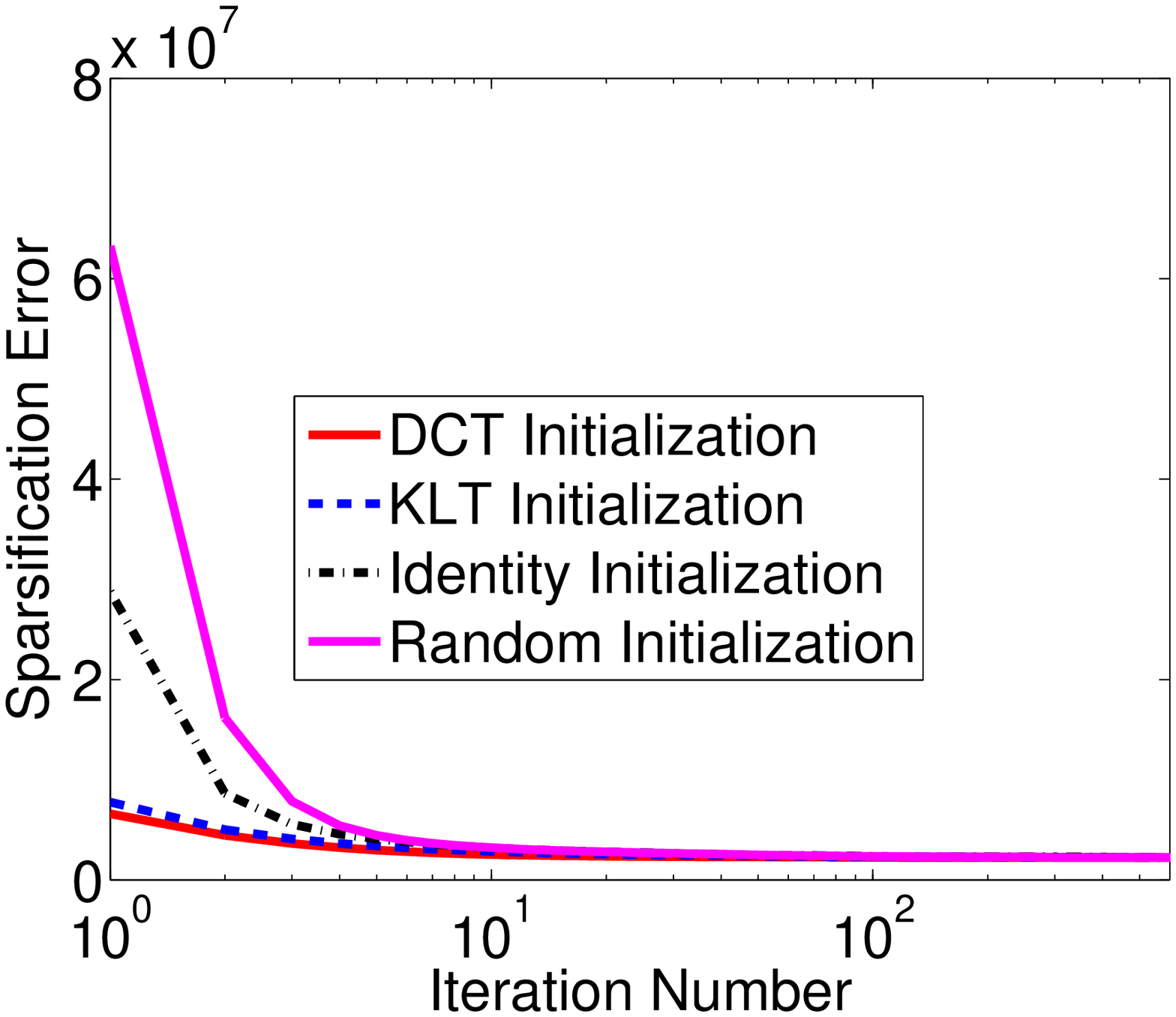}\\
(a) & (b) \\
\end{tabular}
\begin{tabular}{cc}
\includegraphics[height=1.38in]{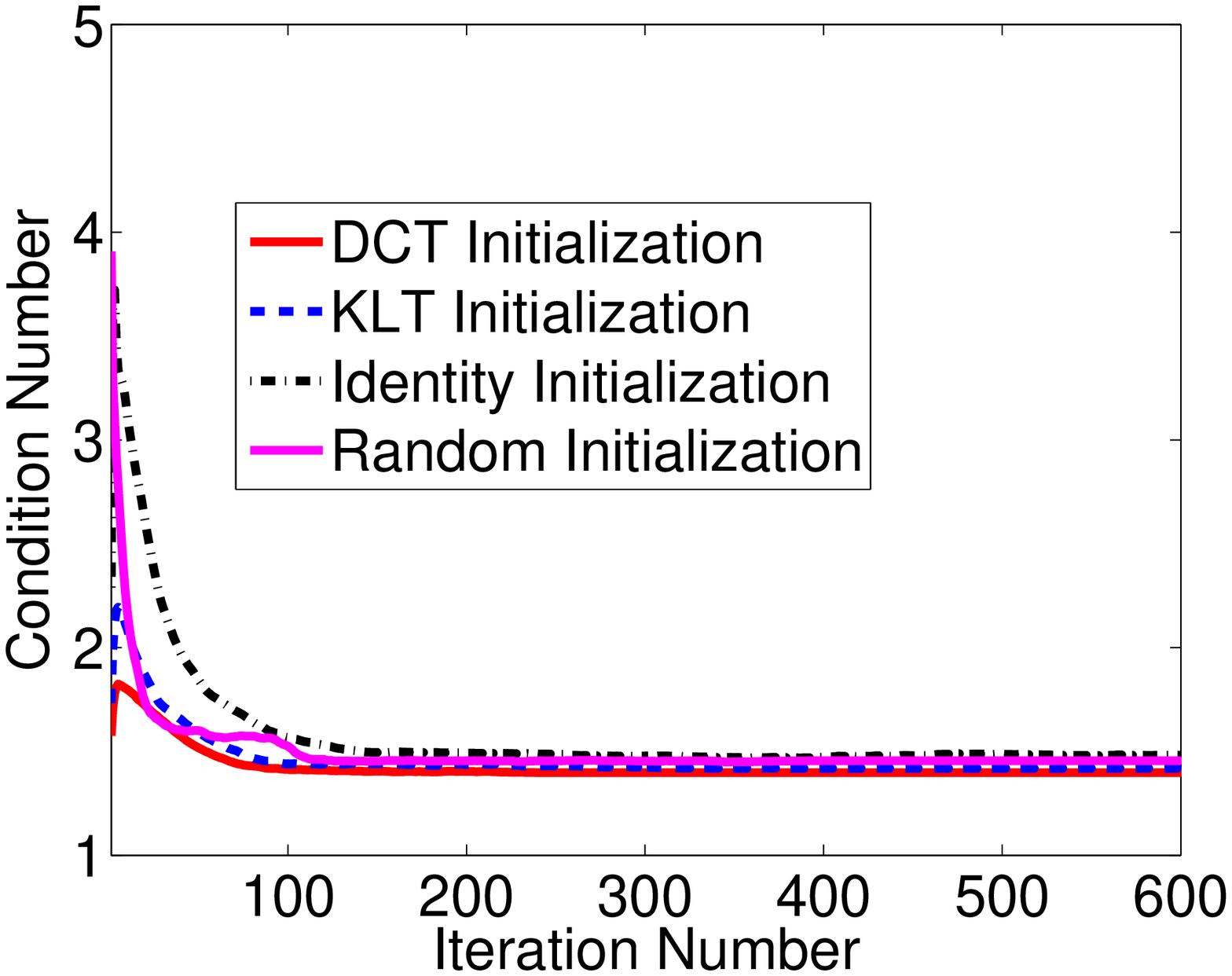}&
\includegraphics[height=1.38in]{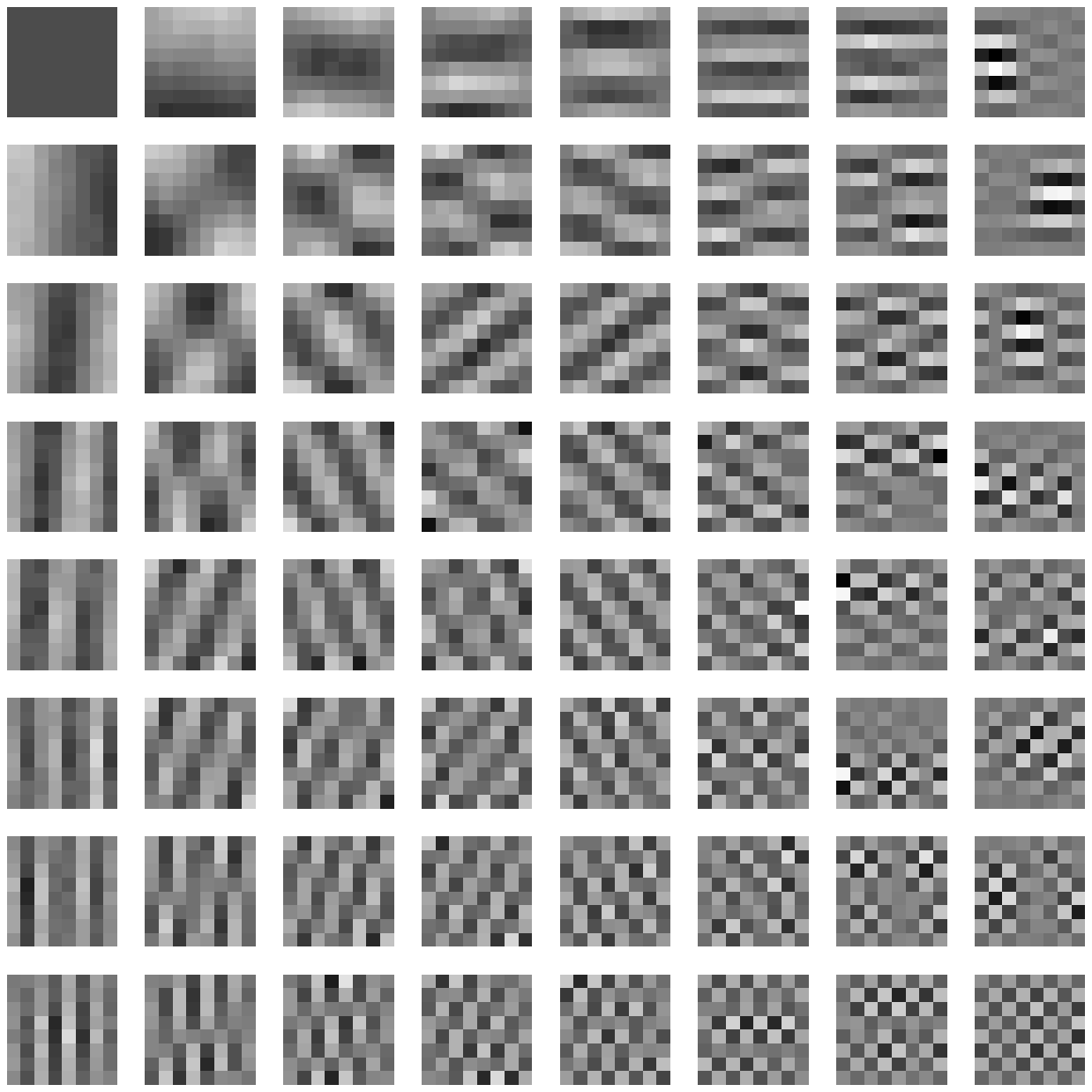}\\
(c) & (d) \\
\end{tabular}
\caption{Effect of different Initializations: (a) Objective function, (b) Sparsification error, (c) Condition number, (d)  Rows of the learnt transform shown as patches for the case of DCT initialization.}
\label{im2}
\end{center}
\end{figure}

We study the convergence behavior of Algorithm A1 for various initializations of $W$. Once $W$ is initialized, the algorithm iterates over the sparse coding and transform update steps (this corresponds to a different ordering of the steps in  Fig. \ref{im5p}). 
We consider four different initializations (initial transforms) for the algorithm. The first is the $64 \times 64$ 2D DCT matrix (obtained as the Kronecker product of two $8 \times 8$ 1D DCT matrices). The second initialization is the Karhunen-Lo\`{e}ve Transform (KLT) (i.e., the inverse of PCA), obtained here by inverting/transposing the left singular matrix of $Y$ \footnote{We did not remove the means of the rows of $Y$ here. However, we obtain almost identical plots in Fig. \ref{im2}, when the learning algorithm is instead initialized with the KLT computed on (row) mean centered data $Y$.}.
The third and fourth initializations are the identity matrix, and a random matrix with i.i.d. Gaussian entries (zero mean and standard deviation 0.2), respectively.


Figure \ref{im2} shows the progress of the algorithm over iterations for the various initializations of $W$. The objective function (Fig. \ref{im2}(a)), sparsification error (Fig. \ref{im2}(b)), and condition number (Fig. \ref{im2}(c)), all converge quickly for our algorithm. The sparsification error decreases over the iterations, as required. Importantly, the final values of the objective (similarly, the sparsification error, and condition number) are nearly identical for all the initializations. This indicates that our learning algorithm is reasonably robust, or insensitive to initialization. Good initializations for $W$ such as the DCT and KLT lead to faster convergence of learning. The learnt transforms also have identical Frobenius norms ($5.14$) for all the initializations. 

Figure \ref{im2}(d) shows the (well-conditioned) transform learnt with the DCT initialization. Each row of the learnt $W$ is displayed as an $8 \times 8$ patch, called the transform atom. The atoms here exhibit frequency and texture-like structures that sparsify the patches of Barbara. Similar to our prior work \cite{sabres}, we observed that the transforms learnt with different initializations, although essentially equivalent in the sense that they produce similar sparsification errors and are similarly scaled and conditioned, appear somewhat different (i.e., they are not related by only row permutations and sign changes). The transforms learnt with different initializations in Fig. \ref{im2} also provide similar recovery PSNRs (that differ by hundredths of a dB) for the Barbara image.



\subsection{Image Representation}

For the second experiment, we learn sparsifying transforms from the $\sqrt{n} \times \sqrt{n}$ (zero mean) non-overlapping patches of the image Barbara at various patch sizes $n$. We study the image representation performance of the proposed algorithm involving closed-form solutions for Problem (P0). We compare the performance of our algorithm to the NLCG-based algorithm \cite{sabres} that solves a version (without the absolute value within the log-determinant) of (P0), and the fixed 2D DCT. The DCT is a popular analytical transform that has been extensively used in compression standards such as JPEG. We set $s=0.17 \times n$ (rounded to nearest integer), and $\lambda_{0}$ is fixed to the same value as in Section \ref{convb} for simplicity. The NLCG-based algorithm is executed with 128 NLCG iterations for each transform update step, and a fixed step size of $10^{-8}$ \cite{sabres}. 



\begin{figure}[!t]
\begin{center}
\begin{tabular}{cc}
\includegraphics[height=1.24in]{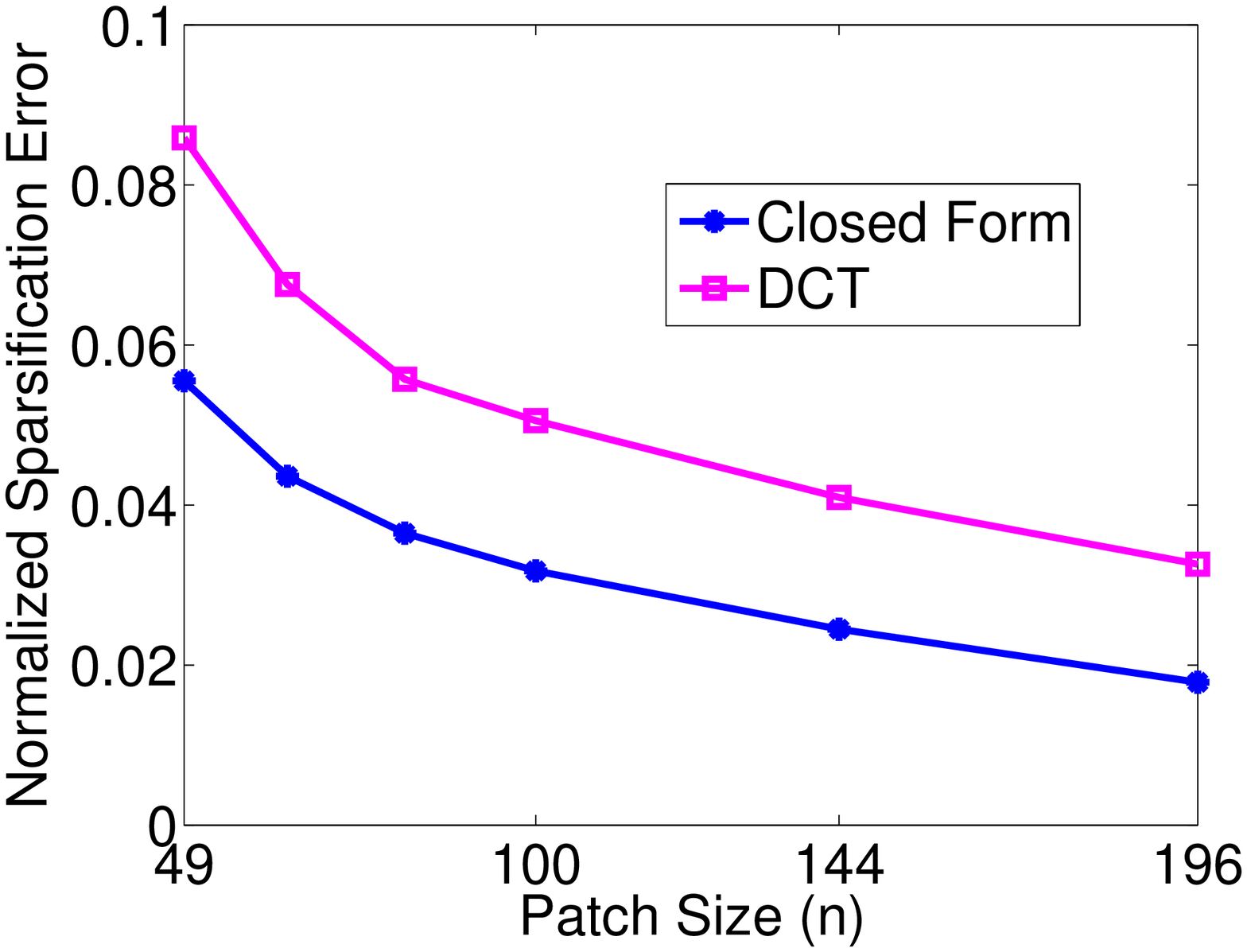}&
\includegraphics[height=1.24in]{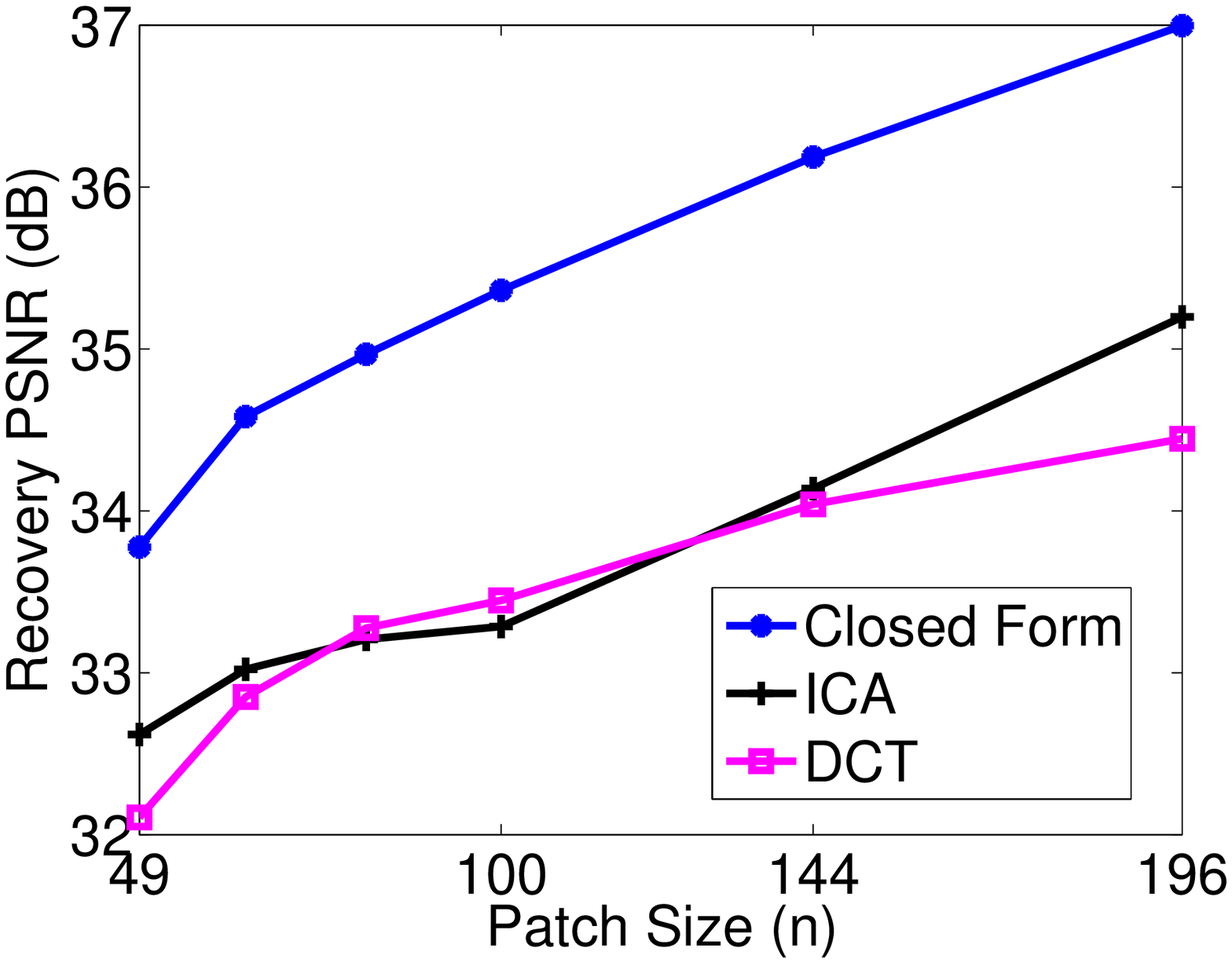}\\
(a) & (b) \\
\end{tabular}
\begin{tabular}{c}
\includegraphics[height=1.24in]{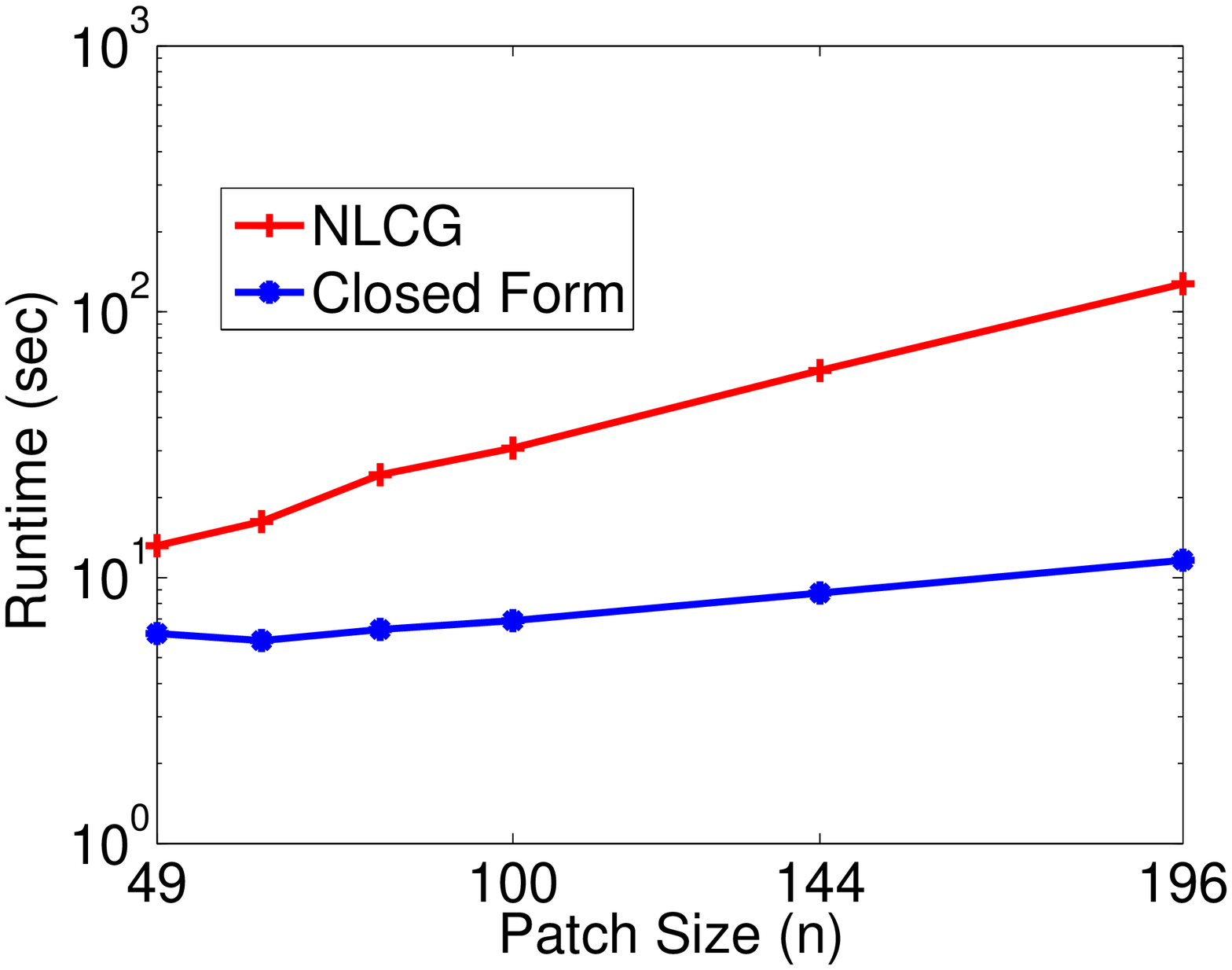}\\
(c) \\
\end{tabular}
\caption{Comparison of NLCG-based transform learning \cite{sabres}, Closed Form transform learning via (P0), DCT, and ICA \cite{fica21} for different patch sizes: (a) Normalized sparsification error, (b) Recovery PSNR, (c) Runtime of transform learning. The plots for the NLCG and Closed Form methods  overlap in (a) and (b). Therefore, we only show the plots for the Closed Form method there.}
\label{im1}
\end{center}
\end{figure}

Figure \ref{im1} plots the normalized sparsification error (Fig. \ref{im1}(a)) and recovery PSNR (Fig. \ref{im1}(b)) metrics for the learnt transforms, and for the patch-based 2D DCT, as a function of patch size. The runtimes of the various transform learning schemes (Fig. \ref{im1}(c)) are also plotted. 



The learnt transforms provide better sparsification and recovery than the analytical DCT at all patch sizes. The gap in performance between the adapted transforms and the fixed DCT also increases with patch size (cf. \cite{doubsp2l} for a similar result and the reasoning).  The learnt transforms in our experiments are all well-conditioned (condition numbers $\approx 1.2-1.6$). Note that the performance gap between the adapted transforms and the DCT can be amplified further at each patch size, by optimal choice of $\lambda_{0}$ (or, optimal choice of condition number \footnote{The recovery PSNR depends on the trade-off between the sparsification error and condition number \cite{sabres, doubsp2l}. For natural images, the recovery PSNR using the learnt transform is typically better at $\lambda$ values corresponding to intermediate conditioning or well-conditioning, rather than unit conditioning, since unit conditioning is too restrictive \cite{sabres}.}).


The performance (normalized sparsification error and recovery PSNR) of the NLCG-based algorithm \cite{sabres} is identical to that of the proposed Algorithm A1 for (P0) involving closed-form solutions. However, the latter is much faster (by 2-11 times) than the NLCG-based algorithm. The actual speedups depend in general, on how $J$ (the number of NLCG iterations) scales with respect to $N/n$.

In yet another comparison, we show in Fig. \ref{im1}(b), the recovery PSNRs obtained by employing Independent Component Analysis (ICA -- a method for blind source separation) \cite{bellsej1, bellsej2, ficppp, fica21, Cardoso2}. Similar to prior work on ICA-based image representation \cite{icacomp21}, we learn an ICA model $A$ (a basis here) using the FastICA algorithm \cite{fica21, ficacose}, to represent the training signals as $Y= A Z$, where the rows of $Z$ correspond to independent sources. Note that the ICA model enforces different properties (e.g., independence) than the transform model.
Once the ICA model is learnt (using default settings in the author's MATLAB implementation \cite{ficacose}), the training signals are sparse coded in the learnt ICA model $A$ \cite{icacomp21} using the orthogonal matching pursuit algorithm \cite{pati}, and the recovery PSNR (defined as in Section \ref{expframew}, but with $W^{-1}X$ replaced by $A\hat{Z}$, where $\hat{Z}$ is the sparse code in the ICA basis) is computed. We found that the $A^{\dagger}$ obtained using the FastICA algorithm provides poor normalized sparsification errors (i.e., it is a bad transform model). Therefore, we only show the recovery PSNRs for ICA. As seen in  Fig. \ref{im1}(b), the proposed transform learning algorithm provides better recovery PSNRs than the ICA approach. This illustrates the superiority of the transform model for sparse representation (compression) of images compared to ICA.
While we used the FastICA algorithm in Fig. \ref{im1}(b), we have also observed similar performance for alternative (but slower) ICA methods \cite{Cardoso2, jade12code}.

Finally, in comparison to synthesis dictionary learning, we have observed that algorithms such as K-SVD  \cite{elad} perform slightly better than
the transform learning Algorithm A1 for the task of image representation. However, the learning and application of synthesis dictionaries also imposes a heavy computational burden (cf. \cite{sabres} for a comparison of the runtimes of synthesis K-SVD and NLCG-based transform learning). Indeed, an important advantage of our transform-based scheme for a compression application (similar to classical approaches involving the DCT or Wavelets), is that the transform can be applied as well as learnt very cheaply.

While we adapted the transform to a specific image (i.e., image-specific transform) in Fig. \ref{im1}, a transform adapted to a variety of images (global transform) also performs well in test images \cite{yblsgg}. Both global and image-specific transforms may hold promise for compression.


\subsection{Image Denoising} \label{imden}


The goal of denoising is to recover an estimate of an image $x \in \mathbb{R}^{P}$ (2D image represented as a vector) from its corrupted measurement $y=x+h$, where $h$ is the noise. We work with $h$ whose entries are i.i.d. Gaussian with zero mean and variance $\sigma^{2}$. We have previously presented a formulation \cite{doubsp2l} for patch-based image denoising using adaptive transforms as follows. 
\begin{align} \label{eq34} 
\nonumber & \hspace{-0.024in}\min_{W, \left \{ x_{i} \right \}, \left \{ \alpha_{i} \right \}} \sum _{i=1}^{N} \left \{  \left \| Wx_{i}-\alpha_{i} \right \|_{2}^{2}  + \lambda_{i}  v(W)  + \tau  \left \| R_{i}\:y - x_{i}\right \|_{2}^{2} \right \}  \\
\nonumber & \;\,\;\,\,\;\; s.t. \,\,\,\, \left \| \alpha_{i} \right \|_{0} \leq s_{i}\; \: \forall \,\,  i \,\,\,\hspace{1.65in}   (\mathrm{P2})
\end{align}
Here, $R_{i} \in \mathbb{R}^{n \times P}$ extracts the $\mathrm{i^{th}}$ patch ($N$ overlapping patches assumed) of the image $y$ as a vector $R_{i}y$. Vector $x_{i}  \in \mathbb{R}^{n }$ denotes a denoised version of $R_{i}y$, and $\alpha_{i} \in \mathbb{R}^{n}$ is a sparse representation of $x_{i}$ in a transform $W$, with an apriori unknown sparsity $s_{i}$. The weight $\tau \propto 1/\sigma$ \cite{elad2, doubsp2l}, and $\lambda_{i}$ is set based on the given noisy data $R_{i} y$ as $ \lambda_{0} \left \| R_{i} y \right \|_{2}^{2}$. The net weighting on $v(W)$ in (P2) is then $\lambda= \sum_{i} \lambda_{i}$.  

We have previously proposed a simple \emph{two-step} iterative algorithm to solve (P2) \cite{doubsp2l}, that also estimates the unknown $s_i$. The algorithm iterates over a transform learning step and a variable sparsity update step (cf. \cite{doubsp2l} for a full description of these steps). We use the proposed alternating transform learning Algorithm A1 (involving closed-form updates) in the transform learning step. Once the denoised patches $x_{i}$ are found, the denoised image $x$ is obtained by averaging the $x_{i}$'s at their respective locations in the image \cite{doubsp2l}.

We now present brief results for our denoising framework employing the proposed efficient closed-form solutions in transform learning. We work with the images Barbara, Cameraman, Couple  \footnote{These three well-known images have been used in our previous work \cite{doubsp2l}.}, and Brain (same as the one in Fig. 1 of \cite{bresai}), and simulate i.i.d. Gaussian noise at 5 different noise levels ($\sigma = 5$, $10$, $15$, $20$, $100$) for each of the images. We compare the denoising results and runtimes obtained by our proposed algorithm with those obtained by the adaptive overcomplete synthesis K-SVD denoising scheme \cite{elad2}.  The Matlab implementation of K-SVD denoising \cite{elad2} available from Michael Elad's website \cite{el2} was used in our comparisons, and we used the built-in parameter settings of that implementation.

We use $11 \times 11$ maximally overlapping image patches for our transform-based scheme. The resulting $121 \times 121$ square transform \footnote{We have previously shown reasonable denoising performance for adapted (using NLCG-based transform learning \cite{sabres}) $64 \times 64$ transforms \cite{doubsp2l}. The denoising performance usually improves when the transform size is increased, but with some degradation in runtime.} has about the same number of free parameters as the $64 \times 256$ overcomplete K-SVD dictionary \cite{elad2, el2}. The settings for the various parameters (not optimized) in our transform-based denoising scheme are listed in Table \ref{tabk1a}. 
 At $\sigma=100$, we set the number of iterations of the two-step denoising algorithm \cite{doubsp2l} to $M'=5$ (lower than the value in  Table \ref{tabk1a}), which also works well, and provides slightly smaller runtimes in denoising.





\begin{table}[t]
\centering
\fontsize{8}{10pt}\selectfont
\begin{tabular}{|c|c|c|c|}
\hline
 Parameter & Value & Parameter & Value \\
\hline
    $ n $           &   121         &   $ N'$            &   $32000$          \\
\hline
    $\lambda_{0}$   &  $0.031$           &      $\tau$         &    $0.01/\sigma$        \\
\hline
         $C$        &      $1.04$      &     $s$            &      $12$         \\
\hline
    $ M'$            &    $11$         &   $  M  $         &  $12$       \\
\hline
\end{tabular}
\caption{The parameter settings for our algorithm: $n$ - number of pixels in a patch, $\lambda_0$ - weight in (P2), $C$ - sets threshold that determines  sparsity levels in the variable sparsity update step \cite{doubsp2l}, $M'$ - number of iterations of the two-step denoising algorithm \cite{doubsp2l}, $N'$ -  training size for the transform learning step (the training patches are chosen uniformly at random from all patches in each denoising iteration) \cite{doubsp2l}, $M$ -  number of iterations in transform learning step, $\tau$ - weight in (P2), $s$ - initial sparsity level for patches \cite{doubsp2l}.}
\label{tabk1a}
\end{table}

\begin{table}[t]
\centering
\fontsize{8}{10pt}\selectfont
\begin{tabular}{|c|c|c|c|c|}
\hline
 Image & $\sigma$ & Noisy PSNR & K-SVD  & Transform \\
\hline
\multirow{5}{*}{Barbara}  &  5  &   34.15             &   38.09             &    38.28                \\ \cline{2-5}
  &  10  &       28.14           &     34.42              &     34.55       \\ \cline{2-5}
  &  15  &        24.59          &      32.34              &    32.39        \\ \cline{2-5}
  &  20  &         22.13         &      30.82              &    30.90        \\ \cline{2-5}
  &  100  &   8.11               &      21.86              &    22.42        \\ 
\hline
\multirow{5}{*}{Cameraman}  &  5  &    34.12            &   37.82             &     37.98               \\ \cline{2-5}
  &  10  &     28.14             &    33.72               &     33.87       \\ \cline{2-5}
  &  15  &       24.60           &     31.50               &     31.65       \\ \cline{2-5}
  &  20  &        22.10          &     29.83               &    29.96        \\ \cline{2-5}
  &  100  &      8.14            &    21.75                &    22.01        \\ 
\hline
\multirow{5}{*}{Brain}  &  5  &    34.14            &     42.14           &       42.74             \\ \cline{2-5}
  &  10  &      28.12            &      38.54             &    38.78        \\ \cline{2-5}
  &  15  &        24.62          &      36.27              &   36.43         \\ \cline{2-5}
  &  20  &        22.09          &      34.70              &   34.71         \\ \cline{2-5}
  &  100  &     8.13             &      24.73              &    24.83        \\ 
\hline
\multirow{5}{*}{Couple}  &  5  &      34.16          &  37.29              &      37.35              \\ \cline{2-5}
  &  10  &      28.11            &     33.48              &    33.67        \\ \cline{2-5}
  &  15  &        24.59          &    31.44                &   31.60         \\ \cline{2-5}
  &  20  &       22.11           &    30.01                &     30.17       \\ \cline{2-5}
  &  100  &      8.13            &    22.58                &    22.60        \\ 
\hline  
\end{tabular}
\caption{PSNR values in decibels for denoising with adaptive transforms, along with the corresponding values for $64 \times 256$ overcomplete K-SVD \cite{elad2}. The PSNR values of the noisy images (denoted as Noisy PSNR) are also shown.}
\label{tabk1b}
\end{table}

Table \ref{tabk1b} lists the denoising PSNRs obtained by our transform-based scheme, along with the PSNRs obtained by K-SVD. The transform-based scheme provides better PSNRs than K-SVD for all the images and noise levels considered. The average PSNR improvement (averaged over all rows of Table \ref{tabk1b}) provided by the transform-based scheme over K-SVD is $0.18$ dB. When the NLCG-based transform learning \cite{sabres} is used in our denoising algorithm, the denoising PSNRs obtained are very similar to the ones shown in Table \ref{tabk1b} for the algorithm involving closed-form updates. However, the latter scheme is faster.



\begin{table}[t]
\centering
\fontsize{8}{10pt}\selectfont
\begin{tabular}{|c|c|c|c|c|c|}
\hline
$\sigma$ &  5 & 10  & 15  & 20 & 100 \\
\hline
Average Speedup &  9.82    &  8.26     &  4.94     &  3.45     &  2.16  \\
\hline
\end{tabular}
\caption{The denoising speedups provided by our transform-based scheme (involving closed-form solutions) over K-SVD \cite{elad2}. The speedups are averaged over the four images at each noise level.}
\label{tabk1c}
\end{table}

We also show the average speedups provided by our transform-based denoising scheme  \footnote{Our MATLAB implementation is not currently optimized for efficiency. Therefore, the speedups here are computed by comparing our unoptimized MATLAB implementation (for transform-based denoising) to the corresponding MATLAB implementation \cite{el2} of K-SVD denoising.} over K-SVD denoising in Table \ref{tabk1c}. For each image and noise level, the ratio of the runtimes of K-SVD denoising and transform denoising (involving closed-form updates) is first computed, and these speedups are averaged over the four images at each noise level. The transform-based scheme is about 10x faster than K-SVD denoising at lower noise levels. Even at very high noise ($\sigma=100$), the transform-based scheme is still computationally cheaper than the K-SVD method.

We observe that the speedup of the transform-based scheme over K-SVD denoising decreases as $\sigma$ increases in Table \ref{tabk1c}. This is mainly because the computational cost of the transform-based scheme is dominated by matrix-vector multiplications (see \cite{doubsp2l} and Section \ref{compco}), and is invariant to the sparsity level $s$. On the other hand, the cost of the K-SVD denoising method is dominated by synthesis sparse coding, which becomes cheaper as the sparsity level decreases. Since sparsity levels in K-SVD denoising are set according to an error threshold criterion (and the error threshold $\propto\sigma^{2}$) \cite{elad2, el2}, they decrease with increasing noise in the K-SVD scheme. For these reasons, the speedup of the transform method over K-SVD is lower at higher noise levels in Table \ref{tabk1c}.


We would like to point out that the actual value of the speedup over K-SVD also depends on the patch size used (by each method). For example, for larger images, a larger patch size would be used to capture image information better. The sparsity level in the synthesis model typically scales as a fraction of the patch size (i.e., $s\propto n$). Therefore, the actual speedup of transform-based denoising over K-SVD at a particular noise level would increase with increasing patch (and image) size -- an effect that is not fully explored here due to limitations of space.

Thus,  here, we have shown the promise of the transform-based denoising scheme (involving closed-form updates in learning) over overcomplete K-SVD denoising. Adaptive transforms provide better denoising, and are faster.  The denoising PSNRs shown for adaptive transforms in Table \ref{tabk1b} become even better at  larger transform sizes, or by optimal choice of parameters \footnote{The parameter settings in Table \ref{tabk1a} (used in all our experiments for simplicity) can be optimized for each noise level, similar to \cite{doubsp2l}.}. We plan to combine transform learning with the state-of-the-art denoising scheme BM3D \cite{dbov} in the near future. Since the BM3D algorithm involves some sparsifying transformations, we conjecture that adapting such transforms could improve the performance of the algorithm.



\section{Conclusions}
\label{sec4}

In this work, we studied the problem formulations for learning well-conditioned square sparsifying transforms. The proposed alternating algorithms for transform learning involve efficient updates. In the limit of $\lambda \to \infty$, the proposed algorithms become orthonormal transform (or orthonormal synthesis dictionary) learning algorithms. Importantly, we provided convergence guarantees for the proposed transform learning schemes. 
We established that our alternating algorithms are globally convergent to the set of local minimizers of the non-convex transform learning problems. Our convergence guarantee does not rely on any restrictive assumptions.
The learnt transforms obtained using our schemes provide better representations than analytical ones such as the DCT for images. In the application of image denoising, our algorithm provides comparable or better performance compared to synthesis K-SVD, while being much faster. 
 Importantly, our learning algorithms, while performing comparably (in sparse image representation or denoising) to our previously proposed learning methods \cite{sabres} involving iterative NLCG in the transform update step, are faster.
We discuss the extension of our transform learning framework to the case of overcomplete (or, tall) transforms elsewhere \cite{saiwen, ovicasp2}.








\appendices

\section{Solution of the Sparse Coding Problem \eqref{bbt5}} \label{appspcodepen}



First, it is easy to see that Problem \eqref{bbt5} can be rewritten as follows
\begin{equation} \label{bbt5apeq2}
\sum _{i=1}^{N}  \sum _{j=1}^{n}\min_{X_{ji}}\: \begin{Bmatrix}
\begin{vmatrix}
 (WY)_{ji} - X_{ji}
\end{vmatrix}^{2} +  \eta_{i}^{2} \, \theta \left ( X _{ji} \right )
\end{Bmatrix}
\end{equation}
where the subscript $ji$ denotes the element on the $j^{\mathrm{th}}$ row and $i^{\mathrm{th}}$ column of a matrix, and
\begin{equation} \label{bbt5apeq3}
 \theta \left ( a \right )=\left\{\begin{matrix}
 0&, \; \mathrm{if} \;\, a = 0 \\
1  &, \; \mathrm{if} \;\, a \neq 0
\end{matrix}\right.
\end{equation}

We now solve the inner minimization problem in \eqref{bbt5apeq2} with respect to $X_{ji}$. This corresponds to the problem
\begin{equation} \label{bbt5apeq4}
\min_{X_{ji}}\: \begin{Bmatrix}
\begin{vmatrix}
 (WY)_{ji} - X_{ji}
\end{vmatrix}^{2} +  \eta_{i}^{2} \, \theta\left ( X _{ji} \right ) 
\end{Bmatrix}
\end{equation}
It is obvious that the optimal $\hat{X}_{ji}= 0$ whenever $(WY)_{ji}=0$. In general, we consider two cases in \eqref{bbt5apeq4}.
First, if the optimal $\hat{X}_{ji}= 0$ in \eqref{bbt5apeq4}, then the corresponding optimal objective value is $(WY)_{ji}^{2}$. If on the other hand, the optimal $\hat{X}_{ji} \neq 0$, then we must have $\hat{X}_{ji}=  (WY)_{ji}$, in order to minimize the quadratic term in \eqref{bbt5apeq4}. In this (second) case, the optimal objective value in \eqref{bbt5apeq4} is $\eta_{i}^{2}$. Comparing the optimal objective values in the two cases above, we conclude that
\begin{equation} \label{bbt5apeq5}
\hat{X}_{ji} =\left\{\begin{matrix}
 0&, \; \mathrm{if} \;\, (WY)_{ji}^{2} < \eta_{i}^{2}\\
 (WY)_{ji}  &, \; \mathrm{if} \;\, (WY)_{ji}^{2} > \eta_{i}^{2}
\end{matrix}\right.
\end{equation}
If $\left | (WY)_{ji} \right | = \eta_{i}$, then the optimal $\hat{X}_{ji}$ in \eqref{bbt5apeq4} can be either $(WY)_{ji} $ or $0$, since both values correspond to the minimum value (i.e., $\eta_{i}^{2}$) of the cost in \eqref{bbt5apeq4}.

Based on the preceding arguments, it is clear that a (particular) solution $\hat{X}$ of \eqref{bbt5} can be obtained as  $\hat{X}_{i} =  \hat{H}_{\eta_{i}}^{1} (W Y_{i})$ $\forall \, i$, where the (hard-thresholding) operator $\hat{H}_{\eta}^{1} (\cdot)$ was defined in Section \ref{algosparsecode1}.

\section{Proof of Proposition \ref{propel33}} \label{orthlimproof}

First, in the sparse coding step of (P0), we solve \eqref{z4} (or \eqref{bbt5}) for $\hat{X}$ with a fixed $W$. Then, the $\hat{X}$ discussed in Section \ref{algo} does not depend on the weight $\lambda$, and it remains unaffected as $\lambda \to \infty$. 

Next, in the transform update step, we solve for $\hat{W}$ in \eqref{z5} with a fixed sparse code $X$. The transform update solution \eqref{tru1} does depend on the weight $\lambda$. For a particular $\lambda$, let us choose the matrix  $L_{\lambda}$  (indexed by $\lambda$) as the positive-definite square root $\left (  YY^{T} + 0.5 \lambda I \right )^{1/2}$. By Proposition \ref{propel1}, the closed-form formula \eqref{tru1} is invariant to the specific choice of this matrix. 
Let us define matrix $M_{\lambda}$ as
\begin{equation} \label{mlam}
M_{\lambda}\triangleq \sqrt{0.5 \lambda} \, L_{\lambda}^{-1} Y X^{T}=\left [ (2/\lambda)YY^{T}  +  I \right ]^{-\frac{1}{2}} Y X^{T}
\end{equation}
and its full SVD as $Q_{\lambda} \tilde{\Sigma}_{\lambda}  R_{\lambda}^{T}$. As $\lambda \to \infty$, by \eqref{mlam}, $M_{\lambda} = Q_{\lambda} \tilde{\Sigma}_{\lambda} R_{\lambda}^{T}$ converges to $ M= Y X^{T}$, and it can be shown (see Appendix \ref{apfg1}) that the accumulation points of  $\left \{ Q_{\lambda} \right \}$ and $\left \{ R_{\lambda} \right \}$ (considering the sequences indexed by $\lambda$, and letting $\lambda \to \infty$) belong to the set of left and right singular matrices of $Y X^{T}$, respectively. Moreover, as $\lambda \to \infty$, the matrix $\tilde{\Sigma}_{\lambda}$ converges to a non-negative $n \times n$ diagonal matrix, which is the matrix of singular values of $YX^{T}$.

On the other hand, using \eqref{mlam} and the SVD of $M_{\lambda}$, \eqref{tru1} can be rewritten as follows
\[\hat{W}_{\lambda}= R_{\lambda} \left[\frac{\tilde{\Sigma}_{\lambda}}{ \lambda }+ \left ( \frac{\tilde{\Sigma}_{\lambda}^{2}}{\lambda^{2} }+  I \right )^{\frac{1}{2}}\right] Q_{\lambda}^{T} \left (  \frac{YY^{T}}{0.5 \lambda}  +  I \right )^{-\frac{1}{2}} \]
 
In the limit of $\lambda \to \infty$, using the aforementioned arguments on the limiting behavior of $\left \{ Q_{\lambda} \right \}$, $ \{ \tilde{\Sigma}_{\lambda} \}$, and $\left \{ R_{\lambda} \right \}$, the  above update formula becomes (or, when $YX^{T}$ has some degenerate singular values, the accumulation point(s) of the above formula assume the following form)
\begin{equation} \label{yfe3}
\hat {W} = \hat{R} \hat{Q}^{T}
\end{equation}
where $\hat{Q}$ and $\hat{R}$ above are the full left and right singular matrices of $Y X^{T}$, respectively. (Note that for $\xi \neq 0.5$, the right hand side of \eqref{yfe3} is simply scaled by the constant $1/\sqrt{2 \xi}$.)
It is clear that the updated transform in \eqref{yfe3} above is orthonormal. 

Importantly, as $\lambda \to \infty$ (with $\xi = 0.5$),  the sparse coding and transform update solutions in (P0) coincide with the corresponding solutions obtained by employing alternating minimization on the orthonormal transform learning Problem \eqref{orthprobav}. 
Specifically, the sparse coding step for Problem \eqref{orthprobav} involves the same aforementioned Problem \eqref{z4}.
Furthermore, using the condition $W^{T}W = I$, it is easy to show that the minimization problem in the transform update step of Problem \eqref{orthprobav} simplifies to the form in \eqref{ortp}.
Problem \eqref{ortp} is of the form of the well-known orthogonal Procrustes problem \cite{procst}.
Therefore, denoting the full SVD of $YX^{T}$ by $U \Sigma V^{T}$, the optimal solution in Problem \eqref{ortp} is given exactly as $\hat{W}=VU^{T}$. 
It is now clear that the solution for $W$ in the orthonormal transform update Problem \eqref{ortp} is identical to the limit shown in \eqref{yfe3}. 

Lastly, the solution to Problem \eqref{ortp} is unique if and only if the singular values of $YX^{T}$ are non-zero. The reasoning for the latter statement is similar to that provided in the proof of Proposition \ref{propel1} (in Section \ref{algo}) for the uniqueness of the transform update solution for Problem (P0).
$\;\;\; \blacksquare$




\section{Limit of a sequence of Singular Value Decompositions} \label{apfg1}

\begin{lem}\label{lemma2s2} \vspace{0.02in}
Consider a sequence $\left \{ M_{k} \right \}$ with $M_{k} \in \mathbb{R}^{n \times n}$, that converges to $M$. For each $k$, let $Q_{k} \Sigma_{k} R_{k}^{T}$ denote a full SVD of $M_{k}$. Then, every accumulation point \footnote{Non-uniqueness of the accumulation point may arise due to the fact that the left and right singular vectors in the singular value decomposition (of $M_{k}$, $M$) are non-unique.} $\left ( Q, \Sigma, R \right )$ of the sequence $\left \{ Q_{k}, \Sigma_{k}, R_{k} \right \}$ is such that $Q \Sigma R^{T}$ is a full SVD of $M$. In particular, $\left \{ \Sigma_{k} \right \}$ converges to $\Sigma$, the $n \times n$ singular value matrix of $M$.
\end{lem}

\hspace{0.1in} \textit{Proof:} 
Consider a convergent subsequence $\left \{ Q_{q_k},  \Sigma_{q_k}, R_{q_k}  \right \}$ of the sequence $\left \{ Q_{k}, \Sigma_{k}, R_{k} \right \}$, that converges to the accumulation point  $\left ( Q, \Sigma, R \right )$. It follows that 
\begin{equation} \label{jazz22}
 \lim_{k \to \infty } M_{q_k} = \lim_{k \to \infty }  Q_{q_{k}}  \Sigma_{q_{k}}  R_{q_{k}}^{T}  =  Q  \Sigma  R^{T}
\end{equation}
Obviously, the subsequence $\left \{ M_{q_k} \right \}$ converges to the same limit $M$ as the (original) sequence $\left \{ M_{k} \right \}$. Therefore, we have
\begin{equation} \label{jazz23}
 M = Q  \Sigma R^{T}
\end{equation}
By the continuity of inner products, the limit of a sequence of orthonormal matrices is orthonormal. Therefore the limits $Q$ and $R$ of the orthonormal subsequences $\left \{ Q_{q_k} \right \}$ and $\left \{R_{q_k}  \right \}$ are themselves orthonormal. Moreover, $\Sigma$, being the limit of a sequence $\left \{ \Sigma_{q_k} \right \}$ of non-negative diagonal matrices (each with decreasing diagonal entries), is also a non-negative diagonal (the limit maintains the decreasing ordering of the diagonal elements) matrix. By these properties and \eqref{jazz23}, it is clear that $Q  \Sigma R^{T}$ is a full SVD of $M$. The preceding arguments also indicate that the accumulation point of $\left \{ \Sigma_{k} \right \}$ is  unique, i.e., $\Sigma$. In other words, $\left \{ \Sigma_{k} \right \}$ converges to $\Sigma$, the singular value matrix of $M$.
 $\;\;\;\; \blacksquare$



\section{Main Convergence Proof}\label{conproof}

Here, we present the proof of convergence for our alternating algorithm for (P0), i.e., proof of Theorem \ref{theorem2}. The proof for Theorem \ref{theorem3} is very similar to that for Theorem \ref{theorem2}. The only difference is that the non-negative barrier function $ \psi (X)$  and the operator $H_{s}(\cdot)$  (in the proof of Theorem \ref{theorem2}) are replaced by the non-negative penalty $\sum _{i=1}^{N} \eta_{i}^{2} \left \| X _{i} \right \|_{0}$ and the operator $\hat{H}_{\eta}^{1} ( \cdot )$, respectively. Hence, for brevity, we only provide a sketch of the proof of Theorem \ref{theorem3}.

We will use the operation $\tilde{H_{s}}(b)$ here to denote the \emph{set} of all optimal projections of $b \in \mathbb{R}^{n}$ onto the $s$-$\ell_{0}$ ball, i.e., $\tilde{H_{s}}(b)$ is the set of all minimizers in the following problem.
\begin{equation} \label{dfdc22}
\tilde{H_{s}}(b) = \underset{x \,:\, \left \| x \right \|_{0} \leq s}{\arg\min}\: \left \| x-b \right \|_{2}^{2}
\end{equation}
Similarly, in the case of Theorem \ref{theorem3}, the operation $\hat{H}_{\eta}(b) $ is defined as a  mapping of a vector $b$ to a set as 
\begin{equation} \label{equ889}
 \left ( \hat{H}_{\eta}(b) \right )_{j}=\left\{\begin{matrix}
 0&, \mathrm{if} \;\, \left | b_{j} \right | < \eta \\
\left \{ b_{j}, \, 0 \right \}  & , \mathrm{if} \;\,  \left | b_{j} \right | = \eta \\ 
b_{j}  & , \mathrm{if} \;\, \left | b_{j} \right | > \eta
\end{matrix}\right.
\end{equation}
 The set $\hat{H}_{\eta}(b) $ is in fact, the set of all optimal solutions to \eqref{bbt5}, when $Y$ is replaced by the vector $b$, and $\eta_{1} = \eta$.




 Theorem \ref{theorem2} is now proved by proving the following properties one-by-one.
\begin{enumerate}[(i)]
\item Convergence of the objective in Algorithm A1.
\item Existence of an accumulation point for the iterate sequence generated by Algorithm A1.
\item All the accumulation points of the iterate sequence are equivalent in terms of their objective value.
\item Every accumulation point of the iterate sequence is a fixed point of the algorithm.
\item Every fixed point of the algorithm is a local minimizer of $g(W, X)$ in the sense of \eqref{dde}.
\end{enumerate}
\vspace{0.05in}


The following shows the convergence of the objective.

\begin{lem}\label{lemma2} \vspace{0.02in}
Let $ \left \{ W^{k}, X^{k} \right \}$ denote the iterate sequence generated by Algorithm A1 with data $Y$ and initial $(W^{0}, X^{0})$. Then, the sequence of objective function values $ \left \{ g(W^{k}, X^{k}) \right \}$ is monotone decreasing, and converges to a finite value $g^{*}=g^{*}\left ( W^{0},X^{0} \right )$.
\end{lem}

\hspace{0.1in} \textit{Proof:}  In the transform update step, we obtain a global minimizer with respect to $W$ in the form of the closed-form analytical solution \eqref{tru1}. Thus, the objective can only decrease in this step, i.e., $g(W^{k+1}, X^{k}) \leq g(W^{k}, X^{k}) $. In the sparse coding step too, we obtain an exact solution for $X$ with fixed $W$ as $\hat{X}_{i} =  H_{s}(W Y_{i})$ $\forall \, i$. Thus, $g(W^{k+1}, X^{k+1}) \leq g(W^{k+1}, X^{k})$. Combining the results for the two steps, we have $g(W^{k+1}, X^{k+1}) \leq g(W^{k}, X^{k})$ for any $k$. 

Now, in Section \ref{sec2}, we stated an explicit lower bound for the function $g(W, X) - \psi (X)$. Since $ \psi (X) \geq 0$, we therefore have that the function $g(W, X)$ is also lower bounded. Since the sequence of objective function values $ \left \{ g(W^{k}, X^{k}) \right \}$ is monotone decreasing and lower bounded, it must converge. $\;\;\;\; \blacksquare$





\begin{lem}\label{lemma3} \vspace{0.02in}
The iterate sequence  $ \left \{ W^{k}, X^{k} \right \}$ generated by Algorithm A1 is bounded, and it has at least one accumulation point.
\end{lem}

\hspace{0.1in} \textit{Proof:} The existence of a convergent subsequence for a bounded sequence is a standard result. Therefore, a bounded sequence has at least one accumulation point. We now prove the boundedness of the iterates. Let us denote $g(W^{k}, X^{k})$ as $g^{k}$ for simplicity. We then have the boundedness of $ \left \{ W^{k}\right \}$ as follows. First, since $g^{k}$ is the sum of $v(W^{k})$, and the non-negative sparsification error and $ \psi (X^{k})$ terms, we have that 
\begin{equation} \label{qw4} 
v(W^{k}) \leq g^{k} \leq g^{0}
\end{equation}
where the second inequality above follows from Lemma \ref{lemma2}. Denoting the singular values of $W^{k}$ by $\beta_{i}$ ($1 \leq i \leq n$), we have that $v(W^{k}) =  \sum_{i=1}^{n}  (\xi \beta_{i}^{2} - \log \, \beta_{i})$.  The function $ \sum_{i=1}^{n}  (\xi \beta_{i}^{2} - \log \, \beta_{i})$, as a function of the singular values $\left \{ \beta_{i} \right \}_{i=1}^{n}$ (all positive) is strictly convex,  and it has bounded lower level sets. (Note that the level sets of a function $f: A \subset \mathbb{R}^{n} \mapsto \mathbb{R}$ (where $A$ is unbounded) are bounded if $\lim_{k \to \infty} f(x^{k}) = + \infty$ whenever $\left \{ x^{k} \right \}\subset A $ and $\lim_{k \to \infty} \left \| x^{k} \right \| = \infty$.) This fact, together with \eqref{qw4} implies that $\left \| W^{k} \right \|_{F} = \sqrt{\sum_{i=1}^{n} \beta_{i}^{2}} \leq c_0$ for a constant $ c_{0}$, that depends on $g^0$. The same bound ($ c_{0}$) works for any $k$.

We also have the following inequalities for sequence  $ \left \{ X^{k} \right \}$.
\begin{equation*} 
 \left \| X^{k} \right \|_{F} - \left \| W^{k} Y \right \|_{F} \leq \left \| W^{k} Y - X^{k} \right \|_{F} \leq \sqrt{g^{k}-v_{0}}
\end{equation*}
The first inequality follows from the triangle inequality and the second inequality follows from the fact that $g^{k}$ is the sum of the  sparsification error and $v(W^{k})$ terms (since $ \psi (X^{k}) =0$), and $v(W^{k}) \geq v_{0}$ ($v_0$ defined in Section \ref{sec2}) \cite{sabres}. By Lemma \ref{lemma2}, $\sqrt{g^{k}-v_{0}} \leq \sqrt{g^{0}-v_{0}}$. Denoting $\sqrt{g^{0}-v_{0}}$ by $c_1$, we have
\begin{equation} \label{qw6}
\left \| X^{k} \right \|_{F}  \leq c_{1} + \left \| W^{k} Y \right \|_{F} \leq c_{1} +  \sigma_{1} \left \| W^{k} \right \|_{F}
\end{equation}
where $\sigma_{1}$ is the largest singular value of the matrix $Y$. The boundedness of $ \left \{ X^{k} \right \}$ then follows from the previously established fact that $\left \| W^{k} \right \|_{F} \leq c_{0}$ . $\;\;\;\; \blacksquare$ 


We now prove some important properties (Lemmas \ref{lemmawea}, \ref{lemma5n},  and \ref{lemmaweb}) satisfied by any accumulation point of the iterate sequence $ \left \{ W^{k}, X^{k} \right \}$ in our algorithm.

\begin{lem}\label{lemmawea} \vspace{0.02in}
Any accumulation point $(W^{*}, X^{*})$ of the iterate sequence  $ \left \{ W^{k}, X^{k} \right \}$ generated by Algorithm A1 satisfies
\begin{equation} \label{zp1l}
 X^{*}_{i} \in \tilde{H_{s}} (W^{*} Y_{i}) \,\,\, \forall \,\, i
\end{equation}
\end{lem}
\hspace{0.1in} \textit{Proof:}
 Let $ \left \{ W^{q_k}, X^{q_k} \right \}$ be a subsequence of the iterate sequence converging to the accumulation point $ (W^{*}, X^{*})$.
It is obvious that $W^{*}$ is non-singular. Otherwise, the objective cannot be monotone decreasing over $ \left \{ W^{q_k}, X^{q_k} \right \}$.

We now have that for each (column) $i$ ($1 \leq i \leq N$),
\begin{equation} \label{zp1}
 X^{*}_{i} = \lim_{k \to \infty } X^{q_k}_{i} = \lim_{k \to \infty } H_{s}(W^{q_{k}} Y_{i})\in \tilde{H_{s}} (W^{*} Y_{i})
\end{equation}
where we have used the fact that when a vector sequence $\left \{ \alpha^{k} \right \} $ converges to $\alpha^{*}$, then the accumulation point of the sequence $\left \{ H_{s}(\alpha^{k}) \right \}$  lies in $\tilde{H_{s}} (\alpha^{*})$ \footnote{Since the mapping $H_{s} (\cdot)$ is discontinuous, the sequence $\left \{ H_{s}(\alpha^{k}) \right \}$ need not converge to $ H_{s}(\alpha^{*})$, even though $\left \{ \alpha^{k} \right \} $ converges to $\alpha^{*}$.} (see proof in Appendix \ref{app3}). $\;\;\;\; \blacksquare$ 



\begin{lem}\label{lemma5n} \vspace{0.02in}
All the accumulation points of the iterate sequence $ \left \{ W^{k}, X^{k} \right \}$ generated by Algorithm A1 with initial $(W^{0}, X^{0})$ correspond to the same objective value. Thus, they are equivalent in that sense. 
\end{lem}

\hspace{0.1in} \textit{Proof:} 
Let $ \left \{ W^{q_k}, X^{q_k} \right \}$ be a subsequence of the iterate sequence converging to an accumulation point $ (W^{*}, X^{*})$. Define a function $g'(W, X) = g(W, X) -  \psi (X)$. Then, for any non-singular $W$, $g'(W, X)$ is continuous in its arguments. Moreover, for the subsequence  $ \left \{ W^{q_k}, X^{q_k} \right \}$ and its accumulation point ($ X^{*}_{i} \in \tilde{H_{s}} (W^{*} Y_{i})$ $\forall$ $i$ by Lemma \ref{lemmawea}), the barrier function $ \psi (X)=0$. Therefore,
\begin{align}
\nonumber \lim_{k \to \infty } g(W^{q_k}, X^{q_k} ) & =\lim_{k \to \infty } g'(W^{q_k}, X^{q_k} ) +\lim_{k \to \infty }  \psi ( X^{q_k} ) \\  & = g'(W^{*}, X^{*}) + 0 = g(W^{*}, X^{*})  \label{bn1}
\end{align}
where we have used the continuity of $g'$ at $(W^{*}, X^{*}) $ (since $W^{*}$ is non-singular). Now, since, by Lemma \ref{lemma2}, the objective converges for Algorithm A1, we have that $\lim_{k \to \infty } g(W^{q_k}, X^{q_k} ) = \lim_{k \to \infty } g(W^{k}, X^{k} ) = g^{*} $. Combining with \eqref{bn1}, we have
\begin{equation} \label{tropi1}
g^{*}= g(W^{*}, X^{*})
\end{equation}
Equation \eqref{tropi1} indicates that any accumulation point $(W^{*}, X^{*})$ of the iterate sequence $ \left \{ W^{k}, X^{k} \right \}$ satisfies $g(W^{*}, X^{*}) = g^{*}$, with $g^{*}$ being the limit of  $ \left \{ g(W^{k}, X^{k}) \right \}$.  $\;\;\;\;\;\; \blacksquare$ 



\begin{lem}\label{lemmaweb} \vspace{0.02in}
Any accumulation point $(W^{*}, X^{*})$ of the iterate sequence  $ \left \{ W^{k}, X^{k} \right \}$ generated by Algorithm A1 satisfies
\begin{equation} \label{tropi3}
W^{*} \in \arg \min_{W} \left \| WY-X^{*} \right \|_{F}^{2}+\lambda \xi \left \| W \right \|_{F}^{2}- \lambda \log \,\left | \mathrm{det \,} W \right |
\end{equation}
\end{lem}
\hspace{0.1in} \textit{Proof:}
 Let $ \left \{ W^{q_k}, X^{q_k} \right \}$ be a subsequence of the iterate sequence converging to the accumulation point $ (W^{*}, X^{*})$. We then have (due to linearity) that 
\begin{equation} \label{jam1}
 \lim_{k \to \infty }  L^{-1} Y  \left ( X^{q_k}  \right )^{T}  = L^{-1} Y  \left ( X^{*}  \right )^{T}
\end{equation}

 Let $ Q^{q_k}  \Sigma^{q_k}  \left ( R^{q_k}  \right )^{T}$ denote the full singular value decomposition of $ L^{-1} Y  \left ( X^{q_k}  \right )^{T}$.
Then, by Lemma \ref{lemma2s2} of Appendix \ref{apfg1}, we have that every accumulation point $\left ( Q^{*}, \Sigma^{*}, R^{*} \right )$ of the sequence $\left \{ Q^{q_k}, \Sigma^{q_k}, R^{q_k} \right \}$ is such that $Q^{*} \Sigma^{*}\left ( R^{*} \right )^{T}$ is a full SVD of $L^{-1} Y  \left ( X^{*}  \right )^{T}$, i.e.,
 \begin{equation} \label{jame1}
Q^{*}  \Sigma^{*} \left ( R^{*} \right )^{T} =  L^{-1} Y  \left ( X^{*}  \right )^{T}
\end{equation}
In particular, $\left \{ \Sigma^{q_k} \right \}$ converges to $\Sigma^{*}$, the full singular value matrix of $L^{-1} Y  \left ( X^{*}  \right )^{T}$. Now, for a convergent subsequence of $\left \{ Q^{q_k}, \Sigma^{q_k}, R^{q_k} \right \}$ (with limit $\left ( Q^{*}, \Sigma^{*}, R^{*} \right )$) , using the closed-form formula \eqref{tru1}, we have
\begin{align}
\nonumber & W^{**}  \triangleq \lim_{k \to \infty } W^{q_{n_k} + 1}   \\  \nonumber & =\lim_{k \to \infty }   \frac{R^{q_{n_k}}}{2} \left(\Sigma^{q_{n_k}} + \left ( \left ( \Sigma^{q_{n_k}} \right )^{2}+2\lambda I \right )^{\frac{1}{2}}\right)\left ( Q^{q_{n_k}} \right )^{T}L^{-1}   \\ & = \frac{R^{*}}{2} \left(\Sigma^{*} + \left ( \left ( \Sigma^{*} \right )^{2}+2\lambda I \right )^{\frac{1}{2}}\right)\left ( Q^{*} \right )^{T}L^{-1} \label{bn3}
\end{align}
where the last equality in \eqref{bn3} follows from the continuity of the square root function, and the fact that $\lambda>0$. Equations \eqref{jame1}, \eqref{bn3}, and \eqref{tru1} imply that
\begin{equation} \label{tropi2}
W^{**} \in \arg \min_{W} \left \| WY-X^{*} \right \|_{F}^{2}+\lambda \xi \left \| W \right \|_{F}^{2}- \lambda \log \,\left | \mathrm{det \,} W \right |
\end{equation}
Now, applying the same arguments used in \eqref{bn1} and \eqref{tropi1} to the sequence $\left \{  W^{q_{n_k} + 1}, X^{q_{n_k} } \right \}$, we get that $g^{*}= g(W^{**}, X^{*})$. Combining with \eqref{tropi1}, we get $g(W^{**}, X^{*}) =  g(W^{*}, X^{*})$, i.e., for a fixed sparse code $X^{*}$, $W^{*}$ achieves the same value of the objective as $W^{**}$. This result together with \eqref{tropi2} proves the required result \eqref{tropi3}.
$\;\;\;\; \blacksquare$ 


Next, we use Lemmas \ref{lemmawea}  and \ref{lemmaweb} to show that any accumulation point of the iterate sequence in Algorithm A1 is a fixed point of the algorithm.  

\begin{lem}\label{lemma3g} \vspace{0.02in}
Any accumulation point of the iterate sequence  $ \left \{ W^{k}, X^{k} \right \}$ generated by Algorithm A1 is a fixed point of the algorithm.
\end{lem}

\hspace{0.1in} \textit{Proof:} Let $ \left \{ W^{q_k}, X^{q_k} \right \}$ be a subsequence of the iterate sequence converging to some accumulation point $ (W^{*}, X^{*})$. Lemmas \ref{lemmawea}  and \ref{lemmaweb} then imply that
\begin{equation} \label{zp1lb}
 X^{*} \in   \arg \min_{X} g(W^{*}, X)                    
\end{equation}
\begin{equation} \label{tropi3lb}
W^{*} \in \arg \min_{W} g(W, X^{*})   
\end{equation}

In order to deal with any non-uniqueness of solutions above, we assume for our algorithm that if a certain iterate $W^{k+1}$ satisfies $ g(W^{k+1}, X^{k} ) = g(W^{k}, X^{k} )$, then we equivalently set $W^{k+1} = W^{k}$. Similarly, if $W^{k+1} = W^{k}$ holds, then we set $X^{k+1} = X^{k}$ 
\footnote{This rule is trivially true, except when applied to say an accumulation point $X^{*}$  (i.e., replace $X^{k}$ by $X^{*}$ in the rule) such that  $X^{*}_{i} \in \tilde{H_{s}}(W^{*}Y_{i})$ $\forall$ $i$, but $X^{*}_{i} \neq H_{s}(W^{*}Y_{i})$ for some $i$.}. Under the preceding assumptions, equations \eqref{tropi3lb} and \eqref{zp1lb} imply that if we feed $(W^{*}, X^{*})$ into our alternating algorithm (as initial estimates), the algorithm stays at $(W^{*}, X^{*})$. In other words, the accumulation point $(W^{*}, X^{*})$ is a fixed point of the algorithm. $\;\;\;\; \blacksquare$ 

Finally, the following Lemma \ref{lemma3c} shows that any accumulation point (i.e., fixed point by Lemma \ref{lemma3g}) of the iterates is a local minimizer. Since the accumulation points are equivalent in terms of their cost, Lemma \ref{lemma3c} implies that they are equally good local minimizers.


We will also need the following simple lemma for the proof of Lemma \ref{lemma3c}.
\begin{lem}\label{lemma9a} \vspace{0.02in}
The function $f(G) = tr(G) -  \log \,\left | \mathrm{det \,} (I + G) \right |$ for $G \in \mathbb{R}^{n \times n}$, has a strict local minimum at $G=0$, i.e., there exists an $\epsilon >0$ such that for $\left \| G \right \|_{F} \leq \epsilon $, we have $f(G) \geq f(0)=0$, with equality attained only at $G=0$.
\end{lem}

\hspace{0.1in} \textit{Proof:} 
The gradient of $f(G)$ (when it exists) is given \cite{sabres} as 
\begin{equation}
\nabla_{G} f(G) = I   -  (I + G)^{-T}
\end{equation}
It is clear that $G=0$ produces a zero (matrix) value for the gradient. Thus, $G=0$ is a stationary point of $f(G)$. The Hessian of $f(G)$ can also be derived \cite{mzb228} as $H = (I + G)^{-T} \otimes (I + G)^{-1}$, where ``$\otimes$" denotes the Kronecker product.  The Hessian is $I_{n^{2}}$ at $G=0$. Since this Hessian is positive definite, it means that $G=0$ is a strict local minimizer of $f(G)$.  The rest of the lemma is trivial. $\;\;\;\;\;\; \blacksquare$



\begin{lem}\label{lemma3c} \vspace{0.02in}
Every fixed point $(W, X)$ of our Algorithm A1 is a minimizer of the objective $g(W, X)$ of Problem (P0), in the sense of \eqref{dde} for sufficiently small $dW$, and $\Delta X$ in the union of the regions R1 and R2 in Theorem \ref{theorem2}. Furthermore,  if $ \left \|  W Y_{i} \right \|_{0} \leq s \, \forall \, i$, then $\Delta X$ can be arbitrary. 
\end{lem}

\hspace{0.1in} \textit{Proof:} It is obvious that $W$ is a global minimizer of the transform update problem \eqref{z5} for fixed sparse code $X$, and it thus provides a  gradient value of $0$ for the objective of \eqref{z5}. Thus, we have (using gradient expressions from \cite{sabres})
\begin{equation} \label{ee10}
2 W Y Y^{T}  - 2 XY^{T} + 2 \lambda \xi W - \lambda W^{-T} = 0 
\end{equation}
Additionally, we also have the following optimal property for the sparse code.
\begin{equation} \label{ee11}
X_{i} \in  \tilde{H_{s}}(WY_{i})  \,\,\, \forall \,\, i
\end{equation}
Now, given such a fixed point $(W, X)$, we consider perturbations $dW \in \mathbb{R}^{n \times n}$, and $\Delta X \in \mathbb{R}^{n \times N}$. We are interested in the relationship between $g(W + dW, X + \Delta X)$ and $g(W, X)$. It suffices to consider \emph{sparsity preserving} $\Delta X$, that is $\Delta X$ such that $X + \Delta X$ has columns that have sparsity $\leq s$. Otherwise the barrier function $ \psi (X+ \Delta X) = + \infty$, and $g(W + dW, X + \Delta X) > g(W, X)$ trivially. Therefore, in the rest of the proof, we only consider sparsity preserving $\Delta X$.

For sparsity preserving $\Delta X \in \mathbb{R}^{n \times N}$, we have
\begin{align} 
\nonumber &  g(W + dW, X + \Delta X) =  \left \| WY - X  + (dW)Y - \Delta X \right \|_{F}^{2}  \\
 & \;\;\; + \lambda \xi \,  \left \| W + dW \right \|_{F}^{2} - \lambda \,  \log \,\left | \mathrm{det \,} (W + dW) \right | \label{ee14}
\end{align}
Expanding the two Frobenius norm terms above using the trace inner product  $\left \langle Q, R \right \rangle \triangleq \mathrm{tr} (Q R^{T})$, and dropping the non-negative terms $\left \| (dW)Y - \Delta X \right \|_F^{2}$ and $\lambda \xi \left \| dW \right \|_F^{2}$, we obtain
\begin{align} 
\nonumber  g(W + dW,  & X  + \Delta X) \geq \left \| WY - X \right \|_{F}^{2} + \lambda \xi \left \| W \right \|_{F}^{2}   \\
\nonumber & + 2 \left \langle  WY - X, (dW) Y - \Delta X \right \rangle + 2 \lambda \xi \left \langle W, dW \right \rangle  \\
 & - \lambda \,  \log \,\left | \mathrm{det \,} (W + dW) \right |  \label{ee15}
\end{align}
Using \eqref{ee10} and the identity $ \log \,\left | \mathrm{det \,} (W + dW) \right | =  \log \,\left | \mathrm{det \,} W \right | +  \log \,\left | \mathrm{det \,} (I + W^{-1} dW) \right |$, equation \eqref{ee15} simplifies to
\begin{align} 
\nonumber & g(W +  dW, X + \Delta X) \geq g(W, X) + \lambda \left \langle W^{-T}, dW  \right \rangle  \\
 &\;\; - 2 \left \langle WY - X, \Delta X \right \rangle - \lambda \,  \log \,\left | \mathrm{det \,} (I + W^{-1} dW) \right |  \label{ee16}
\end{align}
Define $G \triangleq W^{-1} dW$. Then, the terms $ \left \langle W^{-T}, dW  \right \rangle  -  \,  \log \,\left | \mathrm{det \,} (I + W^{-1} dW) \right |$ (appearing in \eqref{ee16} with a scaling $\lambda$) coincide with the function $f(G)$ in Lemma \ref{lemma9a}. Therefore, by Lemma \ref{lemma9a}, we have that there exists an $\epsilon >0$ such that for $\left \| W^{-1} dW \right \|_{F} \leq \epsilon $, we have $  \left \langle W^{-T}, dW  \right \rangle  -  \,  \log \,\left | \mathrm{det \,} (I + W^{-1} dW) \right | \geq 0$, with equality attained here only at $dW=0$. Since $\left \| W^{-1} dW \right \|_{F} \leq  \left \| dW \right \|_{F}/\sigma_{n}$, where $\sigma_n$ is the smallest singular value of $W$, we have that an alternative sufficient condition (for the aforementioned positivity of $f(W^{-1} dW)$) is $\left \| dW \right \|_{F} \leq \epsilon \sigma_{n}$. Assuming that $dW$ lies in this neighborhood, equation \eqref{ee16} becomes
\begin{equation} \label{ee17}
g(W + dW, X + \Delta X) \geq g(W, X)- 2 \left \langle WY - X, \Delta X \right \rangle
\end{equation}
Thus, we have the optimality condition $g(W + dW, X + \Delta X) \geq g(W, X)$ for any $dW \in \mathbb{R}^{n \times n}$ satisfying $\left \| dW \right \|_{F} \leq \epsilon \sigma_{n}$ ($\epsilon$ from  Lemma \ref{lemma9a}), and for any $\Delta X \in \mathbb{R}^{n \times N}$ satisfying $\left \langle WY - X, \Delta X \right \rangle \leq 0$. This result defines Region R1.

We can also define a simple local region $\mathrm{R2} \subseteq \mathbb{R}^{n \times N}$, such that any sparsity preserving $\Delta X$ in the region results in $\left \langle WY - X, \Delta X \right \rangle = 0$. Then,  by \eqref{ee17}, $g(W + dW, X + \Delta X) \geq g(W, X)$  holds for $\Delta X \in \mathrm{R2}$. As we now show, the region R2 includes all $\Delta X \in \mathbb{R}^{n \times N}$ satisfying $\left \| \Delta X \right \|_{\infty} < \min_{i}\left \{ \phi_{s}(W Y_{i}) : \left \| W Y_{i} \right \|_{0}>s \right \}$. In the definition of R2, we need only consider the columns of $WY$ with sparsity greater than $s$. To see why, consider the set $\mathcal{A} \triangleq \left \{ i : \left \| W Y_{i} \right \|_{0}>s \right \}$, and its complement $\mathcal{A}^{c} = \left \{ 1,...,N \right \} \setminus  \mathcal{A}$. Then, we have 
\begin{align}
\nonumber \left \langle WY - X, \Delta X \right \rangle  & = \sum_{i \in \mathcal{A} \cup \mathcal{A}^{c}} \Delta X_{i}^{T}\left ( W Y_{i} - X_{i} \right )\\  
& = \sum_{i \in \mathcal{A}} \Delta X_{i}^{T}\left ( W Y_{i} - X_{i} \right ) \label{bat1}
\end{align}
where we used the fact that $W Y_{i} - X_{i} = 0$, $\forall$ $i \in \mathcal{A}^{c}$. It is now clear that $\left \langle WY - X, \Delta X \right \rangle$ is unaffected by the columns of $WY$ with sparsity $\leq s$. Therefore, these columns do not appear in the definition of R2. Moreover, if $\mathcal{A} = \emptyset$, then $ \left \langle WY - X, \Delta X \right \rangle = 0$ for arbitrary $\Delta X \in \mathbb{R}^{n \times N}$, and thus, $g(W + dW, X + \Delta X) \geq g(W, X)$  holds (by \eqref{ee17}) for arbitrary $\Delta X$. This proves the last statement of the Lemma.

Otherwise, assume $\mathcal{A}\neq \emptyset$, $\Delta X \in \mathrm{R2}$, and recall from \eqref{ee11} that $X_{i} \in  \tilde{H_{s}}(WY_{i})$ $\forall$ $i$. It follows by the definition of R2, that for $i \in \mathcal{A}$, any $X_{i}+ \Delta X_{i}$ with sparsity $\leq s$ will have the same sparsity pattern (non-zero locations) as $X_{i}$, i.e., the corresponding $\Delta X_{i}$ does not have non-zeros outside the support of $X_{i}$. Now, since $X_{i} \in  \tilde{H_{s}}(WY_{i})$, $WY_{i} - X_{i}$ is zero on the support of $X_{i}$, and thus,  $\Delta X_{i}^{T}\left ( W Y_{i} - X_{i} \right ) =0$ for all $i \in \mathcal{A}$. Therefore, by \eqref{bat1}, $\left \langle WY - X, \Delta X \right \rangle = 0$, for any sparsity-preserving $\Delta X$ in R2. 
$\;\;\;\;\;\; \blacksquare$

Note that the proof of Theorem \ref{theorem3} also requires Lemma \ref{lemma3c}, but with the objective $g(W, X)$ replaced by $u(W, X)$.
Appendix \ref{app5} briefly discusses how the proof of the Lemma \ref{lemma3c} is modified for the case of Theorem \ref{theorem3}.


\section{Limit of a Thresholded Sequence} \label{app3}


\begin{lem} \label{lemma2s33hjk} \vspace{0.02in}
Consider a  bounded vector sequence $\left \{ \alpha^{k} \right \} $ with $\alpha^{k} \in \mathbb{R}^{n}$, that converges to $\alpha^{*}$. 
Then, every accumulation point of $\left \{ H_{s}(\alpha^{k})  \right \}$ belongs to the set $\tilde{H_{s}}(\alpha^{*})$. 
\end{lem}

\hspace{0.1in} \textit{Proof:} 
If $\alpha^{*} = 0$, then it is obvious that $\left \{ H_{s}(\alpha^{k})  \right \}$ converges to $\tilde{H_{s}}(\alpha^{*}) = 0$. Therefore, we now only consider the case $\alpha^{*} \neq 0$.

First, let us assume that $\tilde{H_{s}} (\alpha^{*})$ (the set of optimal projections of $\alpha^{*}$ onto the $s$-$\ell_{0}$ ball) is a singleton and $\phi_{s}(\alpha^{*})>0$, so that $\phi_{s}(\alpha^{*})-\phi_{s+1}(\alpha^{*}) >0$. Then, for sufficiently large $k$ ($k \geq k_{0}$), we will have $\left \| \alpha^{k} - \alpha^{*} \right \|_{\infty} < (\phi_{s}(\alpha^{*})-\phi_{s+1}(\alpha^{*}))/2$, and then, $H_{s}(\alpha^{k}) $ has the same support set (non-zero locations) $\Gamma$ as $H_{s}(\alpha^{*}) = \tilde{H_{s}}(\alpha^{*})$. As $k \to \infty$, since $\left \| \alpha_{\Gamma}^{k} - \alpha_{\Gamma}^{*} \right \|_{2} \to 0$ (where the subscript $\Gamma$ indicates that only the elements of the vector corresponding to the support $\Gamma$ are considered), we have that  $\left \| H_{s} (\alpha^{k}) - H_{s} (\alpha^{*}) \right \|_{2} \to 0$. Thus, the sequence $\left \{ H_{s}(\alpha^{k})  \right \}$ converges to $H_{s} (\alpha^{*})$ in this case.


Next, when $ \tilde{H}_{s} (\alpha^{*})$ is a singleton, but $\phi_{s}(\alpha^{*})=0$ (and $\alpha^{*} \neq 0$), let $\gamma$ be the magnitude of the non-zero element of $\alpha^{*}$ of smallest magnitude. Then, for sufficiently large $k$ ($k \geq k_{1}$), we will have $\left \| \alpha^{k} - \alpha^{*} \right \|_{\infty} < \gamma/2$, and then, the support of $H_{s}(\alpha^{*}) = \tilde{H_{s}}(\alpha^{*})$ is contained in the support of $H_{s}(\alpha^{k}) $. Therefore, for $k \geq k_1$, we have
\begin{equation} \label{dvfev34}
\left \| H_{s} (\alpha^{k}) - H_{s} (\alpha^{*}) \right \|_{2} = \sqrt{\left \| \alpha_{\Gamma_1}^{k} - \alpha_{\Gamma_1}^{*} \right \|_{2}^{2} + \left \| \alpha_{\Gamma_2}^{k} \right \|_{2}^{2}}
\end{equation}
where $\Gamma_1$ is the support set of $H_{s} (\alpha^{*}) $, and $\Gamma_2$ (depends on $k$) is the support set of $H_{s}(\alpha^{k})$ excluding $\Gamma_1$. (Note that $\alpha^{*} $ and $H_{s} (\alpha^{*})$ are zero on $\Gamma_2$.)  As $k \to \infty$, since $ \alpha^{k}  \to  \alpha^{*}$, we have that $\left \| \alpha_{\Gamma_1}^{k} - \alpha_{\Gamma_1}^{*} \right \|_{2} \to 0$ and $\left \| \alpha_{\Gamma_2}^{k} \right \|_{2} \to 0$. Combining this with \eqref{dvfev34}, we then have that the sequence $\left \{ H_{s}(\alpha^{k})  \right \}$ converges to $H_{s} (\alpha^{*})$ in this case too.

Finally, when $\tilde{H_{s}}(\alpha^{*})$ is not a singleton (there are ties), it is easy to show that for sufficiently large $k$ ($k \geq k _{2}$), the support of $H_{s}(\alpha^{k}) $ for each $k$ coincides with the support of one of the optimal codes in $\tilde{H_{s}}(\alpha^{*})$. 
In this case, as $k \to \infty$ (or, as $ \alpha^{k}  \to  \alpha^{*}$), the distance between $H_{s}(\alpha^{k}) $ and the set $\tilde{H_{s}}(\alpha^{*})$ converges to $0$. Therefore, the accumulation point(s) of $\left \{ H_{s}(\alpha^{k})  \right \}$ in this case, all belong to the set $\tilde{H_{s}}(\alpha^{*})$. 
$\;\;\;\; \blacksquare$

Specifically, in the case of equation \eqref{zp1}, Lemma \ref{lemma2s33hjk}  implies that $X^{*}_{i} \in \tilde{H_{s}}(W^{*} Y_{i})$.








\section{Modifications to Proof of Lemma \ref{lemma3c} for Theorem \ref{theorem3}}  \label{app5}

The (unconstrained) objective $u(W, X)$ here does not have the barrier function $ \psi (X)$, but instead the penalty $\sum _{i=1}^{N} \eta_{i}^{2} \left \| X _{i} \right \|_{0}$.  Let us consider a fixed point $(W, X)$ of the alternating Algorithm A2 that minimizes this objective. For a perturbation $\Delta X \in \mathbb{R}^{n \times N}$ satisfying $\left \| \Delta X \right \|_{\infty} < \min_{i}\left \{ \eta_{i}/2 \right \}$,  it is easy to see (since $X$ satisfies $X_{i} \in  \hat{H}_{\eta_{i}}(W Y_{i})$ $\forall \, i$) that
\begin{equation}
\sum _{i=1}^{N} \eta_{i}^{2} \left \| X _{i} + \Delta X_{i} \right \|_{0} = \sum _{i=1}^{N} \eta_{i}^{2} \left \| X _{i} \right \|_{0} + \sum _{i=1}^{N} \eta_{i}^{2} \left \| \Delta X _{i}^{c} \right \|_{0}
\end{equation}
where $\Delta X _{i}^{c} \in \mathbb{R}^{n}$ is zero on the support (non-zero locations) of $X_i$, and matches $\Delta X_{i}$ on the complement of the support of $X_{i}$.

Now, upon repeating the steps in the proof of  Lemma \ref{lemma3c} for the case of Theorem \ref{theorem3}, we arrive at the following counterpart of equation \eqref{ee17}.
\begin{align} \label{ee177}
\nonumber u(W + dW, X + \Delta X) \geq & u(W, X)- 2 \left \langle WY - X, \Delta X \right \rangle \\ & \;\; + \sum _{i=1}^{N} \eta_{i}^{2} \left \| \Delta X_{i}^{c} \right \|_{0}
\end{align}
The term $- 2 \left \langle WY - X, \Delta X \right \rangle $ $ + \sum _{i=1}^{N} \eta_{i}^{2} \left \| \Delta X_{i}^{c} \right \|_{0}$  above can be easily shown to be $\geq 0$ for $\Delta X$ satisfying $\left \| \Delta X \right \|_{\infty} < \min_{i}\left \{ \eta_{i}/2 \right \}$. $\;\;\;\;\;\; \blacksquare$

\ifCLASSOPTIONcaptionsoff
  \newpage
\fi



%

\bibliographystyle{./IEEEtran}
\bibliography{./IEEEabrv,./Closedform_v7}




%








\end{document}